\documentclass[journal,twoside,web]{ieeecolor}
\usepackage{generic}
\usepackage{cite}
\usepackage{amsmath,amssymb,amsfonts}
\usepackage{graphicx}
\usepackage{algorithm,algorithmic}
\usepackage{hyperref}
\hypersetup{hidelinks=true}
\usepackage{textcomp}
\usepackage{tabularx}
\usepackage{booktabs}
\usepackage{placeins}
 \usepackage{multirow} 
\def\BibTeX{{\rm B\kern-.05em{\sc i\kern-.025em b}\kern-.08em
    T\kern-.1667em\lower.7ex\hbox{E}\kern-.125emX}}

\begin{document}
\title{CFR-Net: Collaborative Feature Refinement Network for Medical Image Anomaly Detection}

\author{
Zihan Nie,
Muhao Xu,
Wei Feng,
Sijie Niu,
Yi Wan,
Xunbin Wei,
Jianmei Li,
Weiye Song,
and Zongyuan Ge
\thanks{
This work was supported by the Youth Project of the Natural Science Foundation of
Shandong Province, China, under Grant ZR2023QC262; the Natural Science Foundation
of Shandong Province under Grant ZR2022QF017; the Shandong Province Outstanding
Youth Science Fund Project (Overseas) under Grant 2023HWYQ-023; the National
Natural Science Foundation of China under Grant 62205181; the Taishan Scholar
Foundation of Shandong Province under Grant tsqn202211038; and the Key Technology
Research and Development Program of Shandong Province under Grant 2024CXGC010106.
Zihan Nie, Muhao Xu, Yi Wan, Jianmei Li, and Weiye Song are with the
School of Mechanical Engineering, Shandong University, Jinan, China, and the
Key Laboratory of High Efficiency and Clean Mechanical Manufacture,
Shandong University, Ministry of Education, Jinan, China
(e-mail: 202414371@mail.sdu.edu.cn; ujnmhxu@hotmail.com;
wanyi@sdu.edu.cn; lijianmei@sdu.edu.cn; songweiye@sdu.edu.cn).
Zihan Nie is also with Airdoc-Monash Research, Monash University,
Clayton, VIC 3800, Australia.
Wei Feng and Zongyuan Ge are with Monash University,
Clayton, VIC 3800, Australia, and Airdoc-Monash Research,
Monash University, Clayton, VIC 3800, Australia
(e-mail: \{wei.feng, zongyuan.ge\}@monash.edu).
Sijie Niu is with the Shandong Key Laboratory of Ubiquitous Intelligent Computing,
University of Jinan, Jinan, China
(e-mail: sjniu@hotmail.com).
Xunbin Wei is with the Institute of Medical Technology and Cancer Hospital,
Peking University; the Institute of Advanced Clinical Medicine and Biomedical
Engineering Department, Peking University; and the Peking University
International Cancer Institute, Beijing, China
(e-mail: xwei@bjmu.edu.cn).
Jianmei Li and Weiye Song are the corresponding authors.
}
}

\maketitle

\begin{abstract}
Medical image anomaly detection is central to timely diagnosis and clinical decision support, yet abnormal samples are costly to collect because of disease rarity, privacy concerns, and expert workload. This motivates unsupervised learning from normal images, where abnormalities are detected as deviations from learned normal patterns. However, medical anomalies are often subtle, local, and intertwined with normal anatomical variations, which complicates reliable normality modeling. Distillation-based methods support normality modeling by using frozen pretrained teachers as stable feature references, yet mismatches between generic teacher priors and student representations adapted to medical images can produce residuals unrelated to abnormalities in conventional distillation pipelines. To address this limitation, we propose the Collaborative Feature Refinement Network, which learns normality through a coupled process of shared feature conditioning before decoding and cross-space consistency after decoding. Shared feature conditioning performs medical-aware conditioning on teacher and student features under common rules, while cross-space consistency constrains each decoded stream with the complementary encoder representation for reciprocal normal reconstruction. The coupled process is further stabilized by the homework set reorganization strategy, which periodically refreshes normal training subsets. Experiments on six medical image benchmarks show competitive anomaly classification and strong anomaly localization performance.
\end{abstract}


\section{Introduction}
\label{sec:intro}

\IEEEPARstart{M}{edical} image anomaly detection is a core task for timely diagnosis and clinical decision support, aiming to identify image regions or samples that deviate from healthy anatomical patterns \cite{fernando2021deep, aggarwal2021diagnostic}. Such deviations may correspond to clinically important findings, including tumors, microcalcifications, vascular distortions, and other pathological changes \cite{chen2022recent, litjens2017survey}. However, building large-scale abnormal medical datasets remains challenging, as pathological cases can be scarce, privacy restrictions limit data sharing, and reliable annotations require time-consuming expert review \cite{shen2017deep}. These challenges motivate unsupervised anomaly detection from normal images, where abnormal samples are not required during training.

Unsupervised anomaly detection learns normal pattern distributions and identifies abnormalities as deviations from them \cite{hussein2019lung, tschuchnig2021anomaly}. In medical imaging, this normal-only formulation offers a practical route for learning anatomical regularities without requiring representative abnormal samples during training. Nevertheless, reliable normality modeling remains challenging: medical abnormalities are often subtle, local, and intertwined with normal anatomical variations, thereby weakening the separation between normal and abnormal patterns in complex medical backgrounds, where pathological changes may emerge across different spatial scales and orientations \cite{huang2024adapting}.

\begin{figure}[t]
\centerline{\includegraphics[width=\columnwidth, trim={7.7cm 3.7cm 8.7cm 2.6cm}, clip]{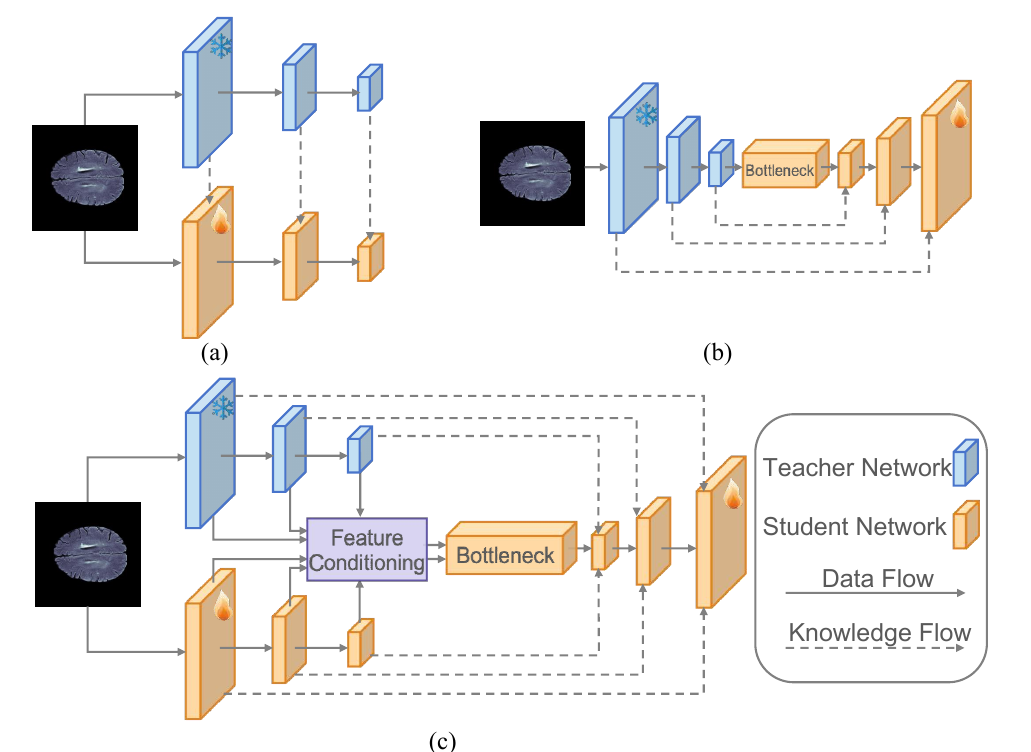}}
\caption{Conceptual comparison of distillation-based anomaly detection frameworks. 
(a) Conventional knowledge distillation performs teacher-to-student matching on normal samples. 
(b) Reverse distillation detects anomalies by comparing teacher references with outputs reconstructed after a bottleneck and decoder. 
(c) CFR-Net introduces shared feature conditioning before decoding and crossed complementary matching after decoding for cross-space normality learning.}
\label{fig:motivation}
\end{figure}

To support normality modeling under these challenges, knowledge distillation provides a natural route by using pretrained networks as stable hierarchical feature references. In teacher and student formulations, as illustrated in Fig.~\ref{fig:motivation}(a), the student mimics the teacher on normal images, and anomalies are detected by enlarged feature discrepancies at test time \cite{bergmann2020uninformed, wang2021student, Salehi_2021_CVPR}. Reverse distillation further strengthens this paradigm by comparing teacher references with outputs reconstructed through a bottleneck and decoder, supporting both image-level scoring and spatial localization, as shown in Fig.~\ref{fig:motivation}(b) \cite{Deng_2022_CVPR}. However, directly transferring this paradigm to medical anomaly detection faces a potential mismatch between frozen teacher priors and trainable student representations. Most pretrained teachers are learned from natural images \cite{wu2022tinyvit}, whose generic priors emphasize edges, textures, and local structures rather than normal anatomical appearances. As student representations adapt to normal medical images, patch-level dispersion, multi-scale contextual variations, and orientation-sensitive structures may be encoded differently from the frozen teacher space, causing conventional reconstruction or direct teacher-student matching to produce high residuals on normal medical patterns rather than true abnormalities.

To address this challenge, we propose the Collaborative Feature Refinement Network (CFR-Net), as illustrated in Fig.~\ref{fig:motivation}(c). CFR-Net couples the Shared Feature Conditioning Module (SFCM) and cross-space consistency into a unified normality learning process: the former performs shared medical-aware conditioning on teacher and student streams, and the latter constrains each decoded stream with the complementary encoder space for crossed teacher-student interaction and reciprocal normal reconstruction. The homework set reorganization strategy further supports this coupled learning process by periodically refreshing normal subsets, reducing dependence on a fixed optimization partition, and improving exposure to normal pattern variations.

Our contributions are summarized as follows:
\begin{itemize}
\item We propose CFR-Net, a cross-space normality learning framework built upon teacher-student distillation, which reformulates distillation-based medical anomaly detection as collaborative normal reconstruction between generic teacher priors and student representations adapted to medical images.

\item We introduce cross-space normality consistency to establish crossed interaction between generic teacher priors and student representations adapted to medical images, enabling reciprocal normal reconstruction through complementary matching.

\item To support cross-space consistency learning, we design SFCM as a shared conditioning module that applies common transformation rules to teacher and student streams before decoding.

\item We validate CFR-Net on six medical image anomaly detection benchmarks, demonstrating competitive anomaly classification and strong anomaly localization performance with comprehensive ablation analyses.
\end{itemize}

\section{Related Work}
\label{sec:related}

\subsection{Feature Representation-based Methods}
Feature representation-based methods detect anomalies by embedding images or local patches into a feature space and measuring their deviations from normal feature distributions. Abnormal regions are typically identified through feature distances, density estimates, or classification scores.

PatchCore \cite{roth2022towards} constructs a memory bank of nominal patch features and uses greedy coreset subsampling to retain representative normal features for nearest-neighbor anomaly scoring. SimpleNet \cite{liu2023simplenet} introduces a feature adapter and a noise-based anomaly feature generator, training a discriminator to separate normal features from synthetic anomalous features. SE-SVDD \cite{hu2021semantic} enhances detection by integrating semantic correlation modules to better capture semantic information. For few-shot scenarios, recent studies have investigated frequency-domain cues and pretrained semantic priors for improving representation robustness under limited normal samples \cite{BAI2024112397, nie2026few}. MOCCA \cite{massoli2021mocca} utilizes a multi-layer one-class classification strategy with multivariate Gaussian distributions to model complex feature relationships.

Flow-based methods detect anomalies by estimating likelihoods in the pretrained feature space. CFLOW-AD \cite{Gudovskiy_2022_WACV} uses conditional normalizing flows to model patch-level feature distributions, while MSFlow \cite{zhou2024msflow} extends likelihood estimation to multi-scale feature distributions. DisAug CLR \cite{sohn2020learning} employs distribution-enhanced contrastive learning to distinguish abnormal representations, while dynamic local augmentation is used to generate negative pairs for self-supervised anomaly detection \cite{yoa2021self}. A recent Mamba-Transformer architecture further integrates local and global feature correlations for multi-class anomaly detection \cite{MA2025113740}.
Despite their effectiveness, these methods largely depend on feature-space separability, which can be affected by domain mismatch and normal anatomical variations in medical images.

\subsection{Reconstruction-based Methods}
Reconstruction-based approaches map input data to latent spaces and reconstruct the inputs, identifying anomalies through input-reconstruction discrepancies.

Bergmann et al. \cite{bergmann2018improving} propose SSIM-based autoencoders that preserve semantic information and image details. Chung et al. \cite{chung2020unsupervised} focus on style-preserving autoencoders to reduce false detections. A multi-scale feature reconstruction network further improves hierarchical anomaly detection by reconstructing features across multiple levels \cite{IQBAL2024112650}. ReContrast \cite{guo2023recontrast} incorporates contrastive learning into feature reconstruction to improve domain-specific anomaly detection.

GAN-based methods such as SCADN \cite{yan2021learning} detect anomalies through partial masking and reconstruction, while AnoSeg \cite{song2021anoseg} combines hard augmentation with channel cascading. Diffusion and score-based generative models have also been explored for reconstruction-based anomaly detection: AnoDDPM \cite{wyatt2022anoddpm} uses denoising diffusion probabilistic models with simplex noise to capture healthy image distributions, while score-based generative modeling has been used to accelerate defect reconstruction through optimized sampling \cite{teng2022unsupervised}. Building on diffusion generative priors, DiAD \cite{he2024diffusion} introduces a diffusion-based framework for multi-class anomaly detection.
However, reconstruction-based methods can be affected by the tension between faithful reconstruction and anomaly preservation, especially when subtle lesions are intertwined with normal anatomical structures.

\subsection{Distillation-based Methods}
Knowledge distillation approaches utilize pretrained teacher networks to guide student networks in learning normal feature representations.

Bergmann et al. \cite{bergmann2020uninformed} pioneered this architecture, using student network uncertainty as anomaly scores. STPM \cite{wang2021student} and MKD \cite{Salehi_2021_CVPR} leverage multi-scale features, with MKD demonstrating that lighter student networks can achieve better performance. RSTPM \cite{yamada2022reconstructed} extends this with additional teacher-student pairs for enhanced anomaly recognition.

RD4AD \cite{Deng_2022_CVPR} introduces multi-scale feature fusion and bottleneck structures, while AST \cite{rudolph2023asymmetric} proposes asymmetric architectures to avoid feature similarity issues. IKD \cite{cao2022informative} addresses overfitting through contextual similarity loss and adaptive hard sample mining. Recent reverse and heterogeneous distillation frameworks further improve anomaly-sensitive feature discrepancies through self-supervised masking or architectural heterogeneity \cite{TONG2023110611, xu2026beyond}. URD \cite{liu2025unlocking} further revisits reverse distillation and improves the representation ability of distillation-based anomaly detection frameworks. These studies demonstrate the effectiveness of distillation-based anomaly detection through feature matching, reverse reconstruction, and asymmetric teacher-student designs. Most existing formulations, however, retain a fixed correspondence between the source and target representation spaces. CFR-Net instead introduces cross-space consistency, where each decoded stream is constrained by the complementary encoder representation, thereby coupling generic teacher priors with student features adapted to normal medical images.

\section{Method}
\subsection{Overall Architecture}
\begin{figure*}[h]
\centerline{\includegraphics[width=0.8\textwidth, trim={6cm 3.5cm 7.5cm 3cm}, clip]{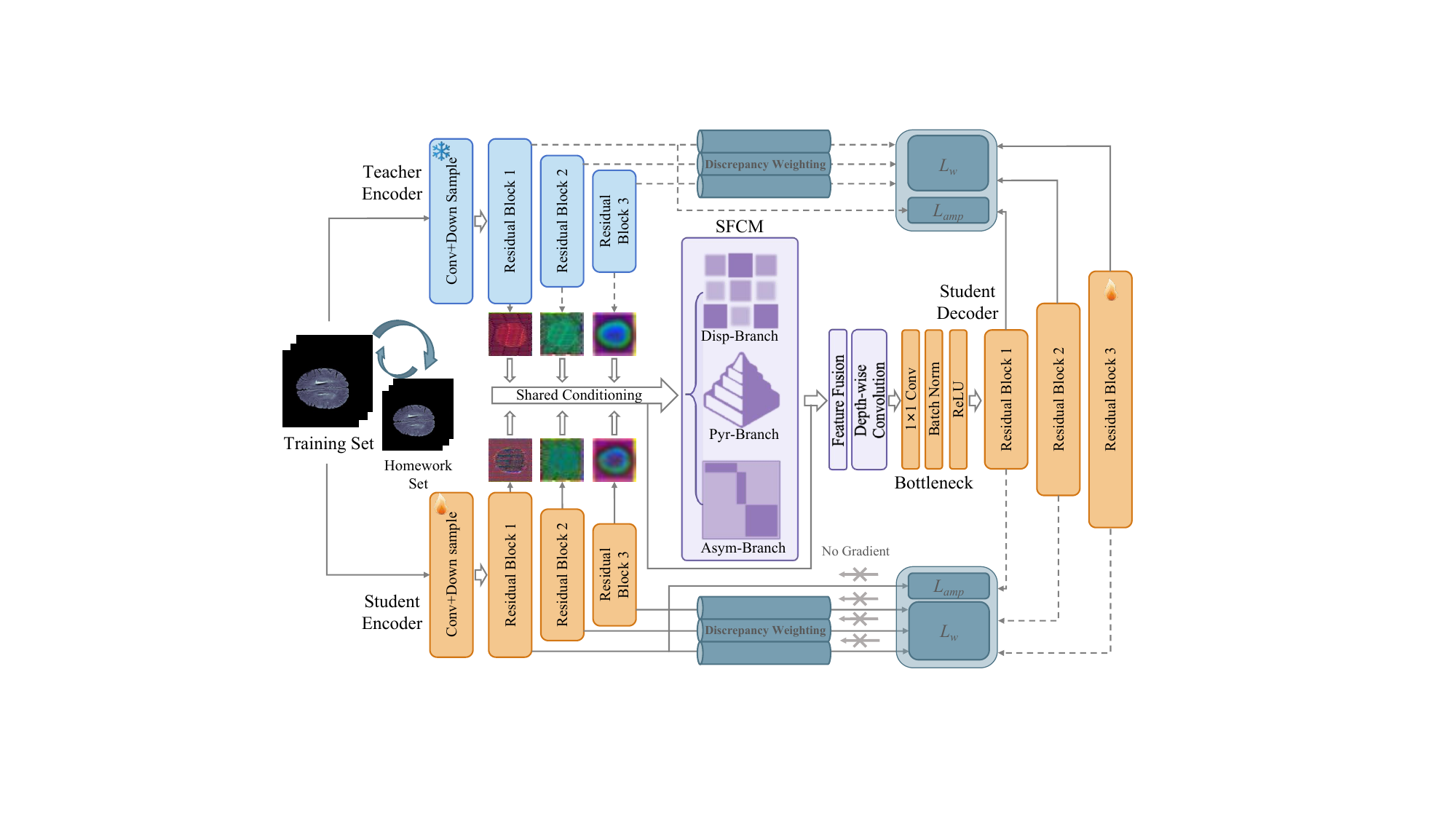}}
\caption{Overall architecture of CFR-Net. A frozen teacher encoder and a trainable student encoder extract hierarchical features from the input image. Features from the teacher encoder and student encoder are processed by SFCM with shared parameters before decoding. The decoded features are optimized with respect to both teacher and student encoder representations through cross-space consistency. During training, the dynamic homework set reorganization strategy periodically refreshes normal subsets for consistency learning.}
\label{fig:framework}
\end{figure*}

CFR-Net is designed for coupled normality learning, where shared feature conditioning before decoding and cross-space consistency after decoding are built upon a teacher-student distillation pipeline, as illustrated in Fig.~\ref{fig:framework}. Given an input image, the frozen teacher encoder and trainable student encoder extract hierarchical feature representations, which are processed by SFCM with shared parameters before decoding. The same SFCM operator is applied to both streams, producing conditioned teacher and student features under common rules. The conditioned features are then decoded and optimized through cross-space consistency, where each decoded stream is constrained by the complementary encoder feature space. 

For an input medical image \(X \in \mathbb{R}^{C \times H \times W}\), preprocessing includes resizing, center cropping, and normalization to ensure consistency with model requirements. The frozen ImageNet-pretrained Wide-ResNet50-2 teacher encoder extracts hierarchical features \(F_{\mathrm{te}}^i \in \mathbb{R}^{C_{\mathrm{T}}^i \times H_{\mathrm{T}}^i \times W_{\mathrm{T}}^i}\), where \(i \in \{1, 2, 3\}\) denotes the layer index from the first three residual block groups, providing generic visual references for consistency learning. In parallel, the trainable student encoder mirrors the teacher structure and extracts corresponding features \(F_{\mathrm{se}}^i\), which provide adaptive representations learned from normal medical images.

For each corresponding feature level, \(F_{\mathrm{te}}^i\) and \(F_{\mathrm{se}}^i\) are processed by SFCM with shared parameters, so that both streams are conditioned by the same operator before cross-space consistency learning. As detailed in Section~\ref{sfcm}, SFCM contains three branches: a Patch-Dispersion Branch for patch-level feature dispersion, a Pyramid Context Branch for contextual aggregation across receptive fields, and an Asymmetric Structure Branch for direction-sensitive structural patterns. The branch outputs are combined with the input encoder features via residual connections.

The conditioned teacher and student features are then passed through depthwise separable convolution, the bottleneck layer, and the student decoder. The bottleneck layer contains three bottleneck units, each combining \(1\times1\) convolution with batch normalization and ReLU activation. After decoding, the outputs are split according to their original feature streams, yielding decoded teacher-stream features \(F_{\mathrm{sd}}^{i,\mathrm{t}}\) and decoded student-stream features \(F_{\mathrm{sd}}^{i,\mathrm{s}}\), where \(\mathrm{t}\) and \(\mathrm{s}\) denote the teacher and student input streams, respectively. Training then compares each decoded stream with the complementary encoder feature space, yielding the cross-space consistency objective defined in Section~\ref{crossspace}. During training, the homework set reorganization strategy periodically repartitions normal samples, exposing the model to refreshed normal variations and reducing dependence on a fixed training split.

\textbf{Inference.}
For inference, the input image is forwarded through the teacher encoder, student encoder, shared SFCM, bottleneck, and decoder to produce decoded teacher-stream and student-stream features. At each feature level, the anomaly response is computed by the same crossed teacher-student comparisons used in consistency learning, while preserving spatial dimensions. Specifically, the level-wise anomaly map is computed as:
\begin{equation}
M^i =
\mathcal{D}_{\mathrm{map}}(F_{\mathrm{te}}^i, F_{\mathrm{sd}}^{i,\mathrm{s}})
+
\mathcal{D}_{\mathrm{map}}(F_{\mathrm{se}}^i, F_{\mathrm{sd}}^{i,\mathrm{t}}),
\end{equation}
where \(\mathcal{D}_{\mathrm{map}}(\cdot,\cdot)\) denotes the spatial cosine discrepancy map, whose scalar counterpart is used for consistency learning in Section~\ref{crossspace}. The final anomaly map is obtained by resizing all level-wise maps to the input resolution using bilinear interpolation and summing them:
\begin{equation}
M = \sum_{i=1}^{3} \operatorname{Up}(M^i),
\end{equation}
where \(\operatorname{Up}(\cdot)\) denotes bilinear interpolation. The image-level anomaly score is computed as the maximum response in \(M\).

\subsection{Cross-Space Consistency Learning}
\label{crossspace}

Following the shared conditioning introduced in the overall architecture, CFR-Net optimizes the decoded teacher-stream and student-stream features through cross-space consistency. The frozen teacher preserves generic pretrained priors, while the trainable student adapts to normal medical images; complementary matching allows each decoded stream to be constrained by both forms of normal representation.

At feature level \(i\), the teacher-origin decoded feature is matched with the student encoder feature, while the student-origin decoded feature is matched with the teacher encoder feature:
\begin{equation}
L_{\mathrm{ts}}^i =
\mathcal{D}(F_{\mathrm{te}}^i, F_{\mathrm{sd}}^{i,\mathrm{s}}),
\quad
L_{\mathrm{st}}^i =
\mathcal{D}(F_{\mathrm{se}}^i, F_{\mathrm{sd}}^{i,\mathrm{t}}).
\end{equation}
During optimization, the student encoder features used as reconstruction references are detached in the consistency objective. This stop-gradient operation blocks the direct gradient path from the matching loss to the student reference features, reducing the risk of trivial feature alignment, while preserving the indirect update path through the bottleneck and decoder.
Here, \(\mathcal{D}(\cdot,\cdot)\) denotes cosine discrepancy between two feature maps:
\begin{equation}
\mathcal{D}(A,B) = 1 -
\frac{\langle \operatorname{vec}(A), \operatorname{vec}(B) \rangle}
{\lVert \operatorname{vec}(A) \rVert_2 \lVert \operatorname{vec}(B) \rVert_2},
\end{equation}
where \(\operatorname{vec}(\cdot)\) denotes feature flattening and \(\langle \cdot,\cdot \rangle\) denotes inner product. The level-wise cross-space consistency loss is then:
\begin{equation}
L_{\mathrm{cross}}^i = L_{\mathrm{ts}}^i + L_{\mathrm{st}}^i.
\end{equation}

\textbf{Hierarchical discrepancy weighting.}
To stabilize cross-space consistency learning across hierarchical feature levels, we adjust the level-wise consistency losses with threshold-based amplification and dispersion-based weighting. A dynamic adaptive factor \(\alpha\) and a minimum loss threshold \(\beta\) are used during training. Both factors are linearly decreased with the training progress \(t \in [0,1]\):
\begin{equation}
\begin{aligned}
\alpha &= \alpha_{\mathrm{initial}} +
(\alpha_{\mathrm{final}} - \alpha_{\mathrm{initial}}) t,\\
\beta &= \beta_{\mathrm{initial}} +
(\beta_{\mathrm{final}} - \beta_{\mathrm{initial}}) t,
\end{aligned}
\end{equation}
where \(\alpha_{\mathrm{initial}}=0.3\), \(\alpha_{\mathrm{final}}=0.2\), \(\beta_{\mathrm{initial}}=0.1\), and \(\beta_{\mathrm{final}}=0.05\), giving a mild decay of the auxiliary adjustment as cross-space consistency learning proceeds.

The level-wise consistency loss is first adjusted by the minimum threshold, so that low-discrepancy levels are not prematurely underemphasized during optimization:
\begin{equation}
L_{\mathrm{amp}}^i =
\begin{cases}
2L_{\mathrm{cross}}^i, & \text{if } L_{\mathrm{cross}}^i < \beta, \\
L_{\mathrm{cross}}^i, & \text{otherwise}.
\end{cases}
\end{equation}

For each feature level, we estimate feature dispersion across the batch to obtain dispersion-based weights. Given a generic feature map \(G^i \in \mathbb{R}^{B \times C_i \times H_i \times W_i}\), where \(B\) is the batch size, its batch-wise feature dispersion is defined as:
\begin{equation}
v(G^i) =
\frac{1}{C_i H_i W_i}
\sum_{c=1}^{C_i}
\sum_{h=1}^{H_i}
\sum_{w=1}^{W_i}
\operatorname{Var}_{b=1}^{B}
\left(G^i_{b,c,h,w}\right).
\end{equation}

The feature-level dispersion for the teacher and student encoder representations is computed as:
\begin{equation}
v_i = \frac{1}{2}
\left(
v(F_{\mathrm{te}}^i) + v(F_{\mathrm{se}}^i)
\right).
\end{equation}

Using the adaptive factor \(\alpha\) defined above, the corresponding dispersion-based weight is computed as:
\begin{equation}
W_i = \exp(\alpha \cdot v_i).
\end{equation}

After normalization across feature levels, the weight for level \(i\) is:
\begin{equation}
\hat{W}_i = \frac{W_i}{\sum_{r=1}^{3} W_r}.
\end{equation}
The normalized weights are applied to the amplified level-wise losses, whereas an unweighted term is retained to preserve the overall consistency constraint:
\begin{equation}
L_{\mathrm{w}} = \sum_{i=1}^{3} \hat{W}_i L_{\mathrm{amp}}^i,
\quad
L_{\mathrm{total}} = L_{\mathrm{w}} + \lambda \sum_{i=1}^{3} L_{\mathrm{amp}}^i,
\end{equation}
where \(\lambda\) is the balancing coefficient.

\subsection{Shared Feature Conditioning Module (SFCM)}
\label{sfcm}
\begin{figure}[t]
\centerline{\includegraphics[width=\columnwidth, trim={6cm 1cm 4.4cm 1cm}, clip]{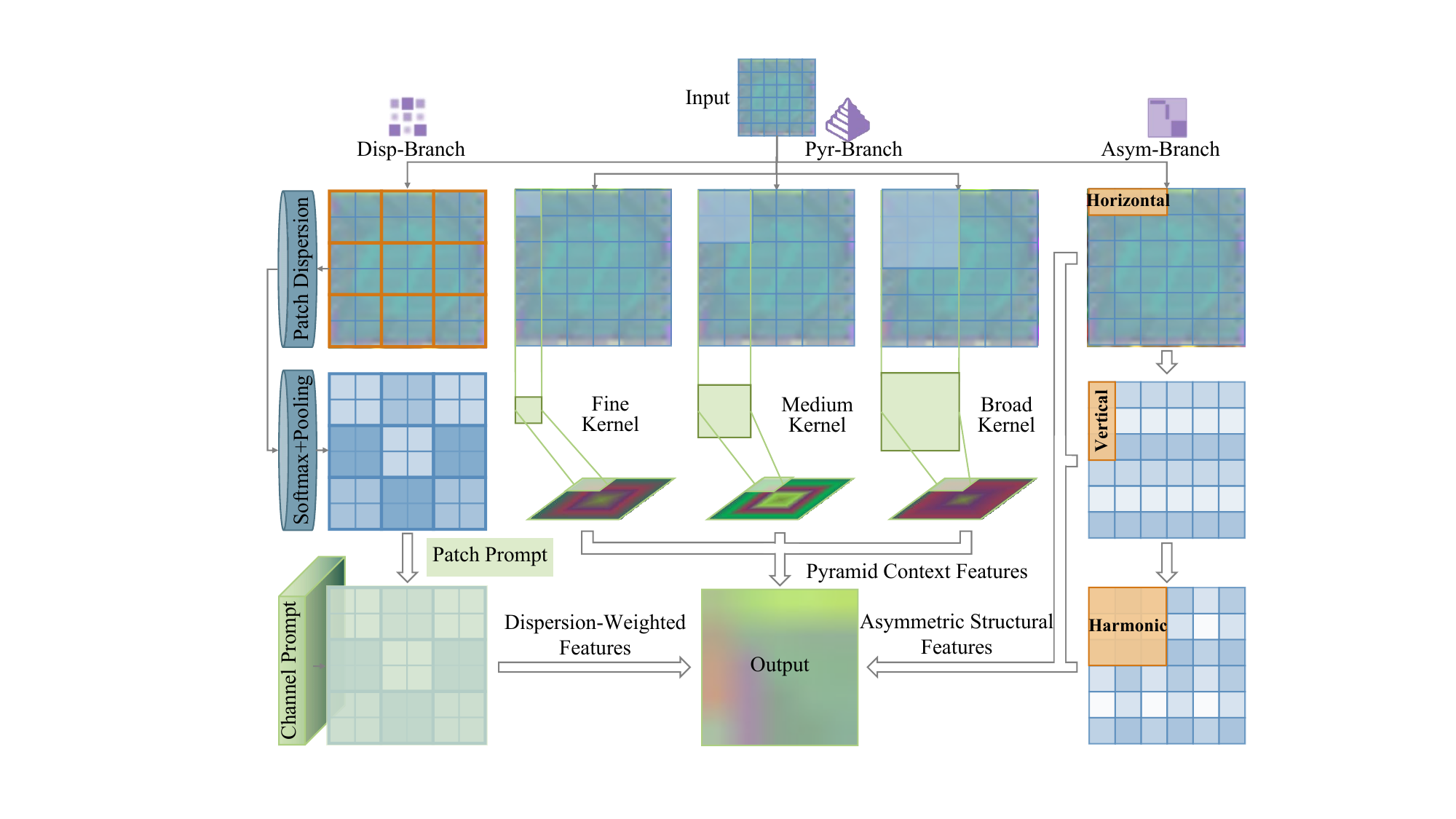}}
\caption{Shared Feature Conditioning Module (SFCM). Disp = Patch-Dispersion Branch, Pyr = Pyramid Context Branch, and Asym = Asymmetric Structure Branch. The module performs shared conditioning on teacher and student encoder features, with branch outputs integrated with the input encoder features through residual connections before decoding.}
\label{fig:SFCM}
\end{figure}

To support the cross-space consistency described above, SFCM is applied level-wise to hierarchical features from the teacher and student encoders, as shown in Fig.~\ref{fig:SFCM}. For each feature level \(i\), \(F_{\mathrm{te}}^i\) and \(F_{\mathrm{se}}^i\) are conditioned by SFCM with shared parameters before decoding. For simplicity, each input encoder feature to SFCM is denoted as \(F_{\mathrm{e}}\), where \(F_{\mathrm{e}} \in \{F_{\mathrm{te}}^i, F_{\mathrm{se}}^i\}\).

\textbf{Patch-Dispersion Branch.}
Patch-level feature dispersion is estimated from local patch statistics. Specifically, for an input encoder feature \(F_{\mathrm{e}}\), the feature map is partitioned into patches of size \(p_h \times p_w\), represented as \(P = \{p_1, p_2, \ldots, p_N\} \in \mathbb{R}^{N \times C_i \times p_h \times p_w}\), where \(N\) represents the number of patches. The feature dispersion of each patch \(V_{p_j}\) is calculated as:
\begin{equation}
V_{p_j} = \frac{1}{C_i p_h p_w} \sum_{k=1}^{C_i p_h p_w} (x_k - \mu_j)^2,
\end{equation}
where \(x_k\) denotes the flattened feature response within patch \(p_j\), and \(\mu_j\) is the mean response of this patch. After computing patch dispersions, softmax generates weights:
\begin{equation}
w_j = \frac{\exp(V_{p_j})}{\sum_{k=1}^N \exp(V_{p_k})}.
\end{equation}
The prompt-modulated patch is computed as:
\begin{equation}
\hat{p}_j =
\mathbf{Q}_{\mathrm{c}}(w_j p_j)
\cdot
\sigma\!\left(\langle \phi(p_j), \mathbf{q}_{\mathrm{p}} \rangle\right),
\end{equation}
where \(\mathbf{Q}_{\mathrm{c}}\) is the learnable channel prompt matrix, \(\mathbf{q}_{\mathrm{p}}\) is the learnable patch prompt, \(\phi(\cdot)\) denotes patch-level feature summarization, and \(\sigma(\langle \phi(p_j), \mathbf{q}_{\mathrm{p}} \rangle)\) produces a patch-level gating coefficient. The output \(F_{\mathrm{disp}}\) is formed by placing each \(\hat{p}_j\) back to its original patch location.

\textbf{Pyramid Context Branch.}
Multi-scale contextual patterns are modeled by applying pyramid convolution with different kernel sizes to the input encoder feature \(F_{\mathrm{e}}\). The output feature map \(F_\ell\) from the \(\ell\)-th pyramid branch is calculated as:
\begin{equation}
F_\ell = \operatorname{Conv}_{k_\ell}(F_{\mathrm{e}}), \quad \ell = 1, 2, \ldots, n,
\end{equation}
where \(k_\ell\) denotes the kernel size of the \(\ell\)-th convolution branch. The final output is obtained by element-wise summation:
\begin{equation}
F_{\mathrm{pyr}} = \sum_{\ell=1}^{n} F_\ell.
\end{equation}

\textbf{Asymmetric Structure Branch.}
Orientation-sensitive structural patterns are modeled through factorized sequential convolutions. Horizontal, vertical, and harmonic kernels are applied in order, where each stage takes the output of the previous stage as input:
\begin{equation}
F_{\mathrm{hori}} = \operatorname{Conv}_{\mathrm{hori}}(F_{\mathrm{e}}),
\end{equation}
\begin{equation}
F_{\mathrm{vert}} = \operatorname{Conv}_{\mathrm{vert}}(F_{\mathrm{hori}}),
\end{equation}
\begin{equation}
F_{\mathrm{harm}} = \operatorname{Conv}_{\mathrm{harm}}(F_{\mathrm{vert}}).
\end{equation}
The branch output is obtained by stage-wise summation:
\begin{equation}
F_{\mathrm{asm}} = F_{\mathrm{hori}} + F_{\mathrm{vert}} + F_{\mathrm{harm}}.
\end{equation}

The features from all three branches are fused with the original encoder feature through element-wise addition:
\begin{equation}
F_{\mathrm{fused}} = F_{\mathrm{e}} + F_{\mathrm{disp}} + F_{\mathrm{pyr}} + F_{\mathrm{asm}}.
\end{equation}

The fused features \(F_{\mathrm{fused}}\) then undergo depthwise separable convolution and are passed to the bottleneck and the student decoder, producing decoded features from the teacher stream \(F_{\mathrm{sd}}^{i,\mathrm{t}}\) and decoded features from the student stream \(F_{\mathrm{sd}}^{i,\mathrm{s}}\) according to their original encoder streams.

\subsection{Homework Set Reorganization Strategy}

CFR-Net adopts a homework set reorganization strategy to diversify the normal samples encountered during optimization. Given the normal training set $\mathcal{X}_{\mathrm{tr}}$, samples are randomly repartitioned at the beginning of each training cycle into a training subset $\mathcal{X}_{\mathrm{sub}}$ and a homework subset $\mathcal{X}_{\mathrm{hw}}$. The model is first optimized on $\mathcal{X}_{\mathrm{sub}}$ for five subset epochs and is then optimized for one additional pass over $\mathcal{X}_{\mathrm{hw}}$. The two subsets are subsequently reshuffled for the next cycle, so that their roles change periodically throughout training.

In our implementation, the homework ratio is set to $\rho=0.5$, and training consists of 20 reorganization cycles. Each cycle contains five optimization epochs over $\mathcal{X}_{\mathrm{sub}}$, followed by one optimization pass over $\mathcal{X}_{\mathrm{hw}}$. The teacher encoder remains frozen, while SFCM, the student encoder, bottleneck, and decoder are updated. With $\rho=0.5$, this schedule does not increase the overall sample exposure relative to conventional full-set training; instead, it reorganizes how normal samples are presented during optimization. The procedure is summarized in Algorithm~\ref{alg:homework}.

\begin{algorithm}[h]
\caption{Homework Set Reorganization Strategy}
\label{alg:homework}
\begin{algorithmic}[1]
\REQUIRE Normal training set $\mathcal{X}_{\mathrm{tr}}$, homework ratio $\rho$, number of training cycles $C$
\ENSURE Optimized CFR-Net
\FOR{cycle $=1$ to $C$}
    \STATE Randomly repartition $\mathcal{X}_{\mathrm{tr}}$ into $\mathcal{X}_{\mathrm{sub}}$ and $\mathcal{X}_{\mathrm{hw}}$ with homework ratio $\rho$
    \FOR{$m=1$ to $5$}
        \STATE Optimize CFR-Net for one epoch on $\mathcal{X}_{\mathrm{sub}}$ using $L_{\mathrm{total}}$
    \ENDFOR
    \STATE Optimize CFR-Net for one pass on $\mathcal{X}_{\mathrm{hw}}$ using $L_{\mathrm{total}}$
\ENDFOR
\end{algorithmic}
\end{algorithm}

\section{Experimental Setup}

\subsection{Datasets}
We evaluate CFR-Net on six medical image anomaly detection benchmarks covering MRI, CT, retinal OCT, fundus imaging, and histopathology. For BrainMRI, LiverCT, RESC, OCT17, and HIS, we follow the dataset construction and train-test splits defined by the BMAD benchmark~\cite{bao2024bmad}. All methods are trained using normal images only, and abnormal samples are used only for testing or evaluation.

The BrainMRI dataset is constructed from BraTS2021~\cite{baid2021rsnaasnrmiccaibrats2021benchmark,Bakas2017,6975210}. Following the 2D slice-based anomaly detection setting, we use FLAIR volumes and select slices from depths 60 to 100 to account for anatomical variations across different brain depths. Each extracted slice is saved as a PNG image with an original size of $240\times240$. The dataset contains 7,500 normal training slices and 3,715 test slices with a balanced ratio between normal and anomalous samples, and provides segmentation masks for anomaly localization.

The LiverCT dataset is built from BTCV~\cite{landman2015miccai} and LiTS~\cite{BILIC2023102680}. The anomaly-free BTCV set, consisting of 50 abdominal 3D CT scans, is used to construct the normal training set, while 131 abdominal 3D CT scans from LiTS are used for testing. Hounsfield Unit values are converted to grayscale images under an abdominal window, and the scans are cropped into 2D axial slices. The resulting dataset contains 1,452 normal training slices and 1,493 test slices, with segmentation masks available for localization evaluation.

For retinal OCT analysis, we use RESC~\cite{HU2019216} and OCT17~\cite{kermany2018identifying}, which provide complementary evaluation settings for localization and classification. RESC contains retinal OCT images with pixel-level segmentation annotations for retinal edema, and is used for anomaly localization. It includes 4,297 normal training images and 1,805 test images, with an original image size of $(512\times1024)$. OCT17 is used for anomaly classification and contains retinal OCT images categorized into three anomaly types. It includes 26,315 normal training images and 968 test images, with an original image size of $(512\times496)$.

The APTOS dataset is constructed from the official APTOS 2019 Blindness Detection Challenge~\cite{aptos2019-blindness-detection}, following the split used in prior anomaly detection work~\cite{guo2023recontrast}. It contains 3,662 color fundus images annotated with diabetic retinopathy grades from 0 to 4. Grade 0 images are treated as normal, while grades 1 to 4 are treated as anomalous. Among 1,805 normal images and 1,857 anomalous images, 1,000 normal images are randomly selected for training, and the remaining 2,662 images are used for testing.

The HIS dataset is based on Camelyon16~\cite{10.1001/jama.2017.14585}, which contains hematoxylin and eosin stained lymph node whole-slide images from breast cancer patients. We use 5,088 randomly extracted normal patches for training. The test set contains 1,003 normal patches and 997 abnormal patches, and is used for image-level anomaly classification.

\subsection{Benchmark Protocol}
For BrainMRI, LiverCT, RESC, OCT17, and HIS, we follow the dataset construction, train-test splits, and evaluation protocol defined by the BMAD benchmark~\cite{bao2024bmad}. The benchmark results of CFLOW-AD~\cite{Gudovskiy_2022_WACV}, RD4AD~\cite{Deng_2022_CVPR}, PatchCore~\cite{roth2022towards}, and MKD~\cite{Salehi_2021_CVPR} are also obtained directly from BMAD under these established settings. For APTOS, we follow the same split as prior work~\cite{guo2023recontrast} and reproduce all compared methods under our experimental setup. Methods without available BMAD results on the evaluated datasets, including SimpleNet~\cite{liu2023simplenet}, DiAD~\cite{he2024diffusion}, MSFlow~\cite{zhou2024msflow}, and URD~\cite{liu2025unlocking}, are run with their default settings, while CFR-Net uses a single configuration consistently across all six datasets.

\begin{table*}[htbp]
  \centering
  \small
  \setlength{\tabcolsep}{3pt}
  \caption{Quantitative comparison with different anomaly classification and localization methods. Results are reported as AUROC (\%). AC and AL denote anomaly classification and anomaly localization, respectively. The best result is in \textbf{bold}, and the second-best result is \underline{underlined}.}
  \begin{tabularx}{\textwidth}{@{}c c *{9}{>{\centering\arraybackslash}X}@{}}
    \toprule
    \multirow{2}{*}[-2pt]{Method} & \multirow{2}{*}[-2pt]{Source} & HIS & OCT17 & APTOS & \multicolumn{2}{c}{BrainMRI} & \multicolumn{2}{c}{LiverCT} & \multicolumn{2}{c}{RESC} \\
    \cmidrule(r){3-3} \cmidrule(lr){4-4} \cmidrule(lr){5-5} \cmidrule(lr){6-7} \cmidrule(lr){8-9} \cmidrule(l){10-11}
    & & AC & AC & AC & AC & AL & AC & AL & AC & AL \\
    \midrule
    CFLOW-AD \cite{Gudovskiy_2022_WACV} & WACV2022 & 54.54 & 85.43 & 94.21 & 73.97 & 93.52 & 49.93 & 92.78 & 74.43 & 93.75 \\
    RD4AD \cite{Deng_2022_CVPR} & CVPR2022 & 66.59 & 97.24 & 92.43 & \underline{89.38} & 96.54 & 60.02 & 95.86 & 87.53 & 96.17 \\
    PatchCore \cite{roth2022towards} & CVPR2022 & 69.34 & \underline{98.56} & 90.45 & \textbf{91.55} & \underline{96.97} & 60.40 & 96.58 & 91.50 & 96.39 \\
    MKD \cite{Salehi_2021_CVPR} & CVPR2022 & \textbf{77.74} & 96.62 & 84.94 & 81.38 & 89.54 & 60.39 & 96.14 & 88.97 & 86.60 \\
    SimpleNet \cite{liu2023simplenet} & CVPR2023 & 56.26 & 98.31 & 92.80 & 84.35 & 93.22 & \underline{72.49} & 97.39 & 86.74 & 91.65 \\
    DiAD \cite{he2024diffusion} & AAAI2024 & 67.26 & 98.54 & 73.29 & 82.36 & 96.08 & 58.17 & 96.84 & 87.77 & 94.01 \\
    MSFlow \cite{zhou2024msflow} & TNNLS2024 & 55.08 & 96.81 & \underline{94.90} & 88.73 & 95.91 & 68.15 & 96.43 & 91.56 & 94.86 \\
    URD \cite{liu2025unlocking} & AAAI2025 & 67.30 & 98.04 & 93.91 & 77.30 & 96.73 & 72.08 & \underline{97.86} & \underline{92.45} & \underline{96.74} \\
    CFR-Net & Ours & \underline{70.62} & \textbf{99.63} & \textbf{96.07} & 88.42 & \textbf{98.77} & \textbf{74.62} & \textbf{98.48} & \textbf{93.99} & \textbf{98.57} \\
    \bottomrule
  \end{tabularx}
  \label{tab:comparison}
\end{table*}

\begin{figure*}[!h]
\centerline{\includegraphics[width=0.9\textwidth, trim={0cm 0.3cm 0cm 0cm}, clip]{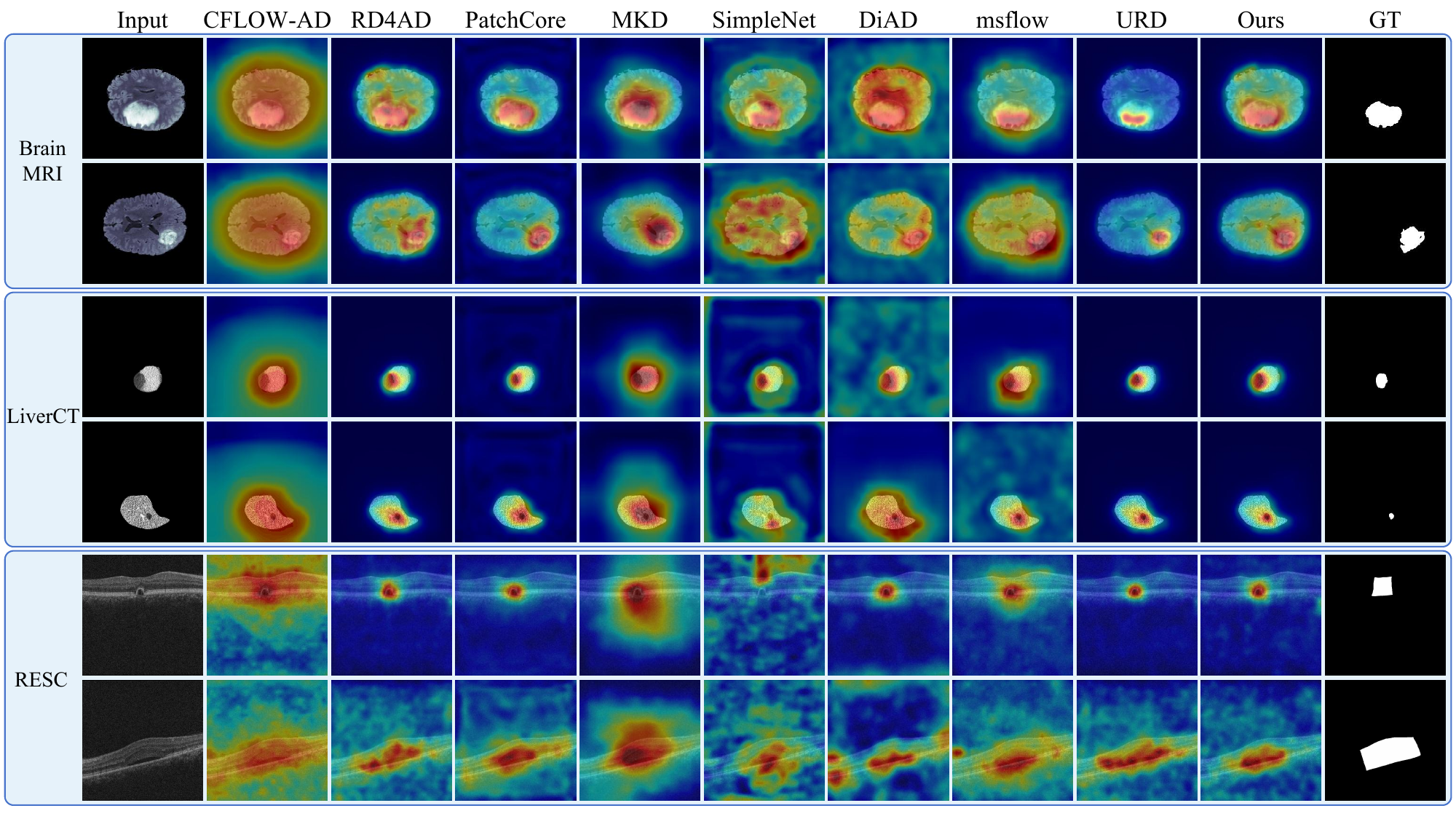}}
\caption{Qualitative comparison of anomaly localization maps. From left to right: input images, predictions from CFLOW-AD, RD4AD, PatchCore, MKD, SimpleNet, DiAD, MSFlow, URD, our method, and ground truth annotations.}
\label{fig:localization}
\end{figure*}

\subsection{Implementation Details}
The experiments were conducted on an NVIDIA A100 GPU with 40 GB memory. All input images were resized to \(256\times256\) pixels and normalized. The implementation used Python 3.8 and PyTorch 1.12.

For optimization, we used the AdamW optimizer with a learning rate of \(2 \times 10^{-3}\), weight decay of \(1 \times 10^{-5}\), and batch size of 32. The teacher encoder leveraged Wide-ResNet50-2 pretrained on ImageNet, whereas student components were randomly initialized. Training consists of 20 reorganization cycles. Each cycle contains five epochs over the training subset followed by one pass over the homework subset. Hyperparameters were set as: \(\alpha_{\mathrm{initial}}=0.3\), \(\alpha_{\mathrm{final}}=0.2\), \(\beta_{\mathrm{initial}}=0.1\), \(\beta_{\mathrm{final}}=0.05\), and \(\lambda=0.1\). SFCM used a patch size of \(4\times4\), with pyramid kernels of \(3\times3\), \(5\times5\), and \(7\times7\) in the Pyramid Context Branch, and asymmetric kernels of \(1\times3\), \(3\times1\), and \(3\times3\) for horizontal, vertical, and harmonic convolutions in the Asymmetric Structure Branch, respectively.

\section{Results and Discussion}
\subsection{Comparison on Anomaly Classification and Localization}

We employ the Area Under the Receiver Operating Characteristic Curve (AUROC) to assess model performance, with higher values indicating better discrimination between normal and anomalous samples.

As shown in Table~\ref{tab:comparison}, CFR-Net achieves leading performance in seven out of nine evaluation tasks, with a consistent advantage in anomaly localization. It obtains the highest AUROC for anomaly localization on BrainMRI, LiverCT, and RESC, suggesting that shared feature conditioning before decoding introduces medical-aware local, contextual, and structural modulation to make teacher and student streams more compatible for collaborative normality learning, thereby reducing misleading discrepancies caused by domain mismatch. Based on these conditioned representations, cross-space consistency further preserves spatially meaningful anomaly responses through complementary teacher-student matching.

\begin{figure*}[h!]
\centerline{\includegraphics[width=0.9\textwidth, trim={0cm 9cm 0cm 0cm}, clip]{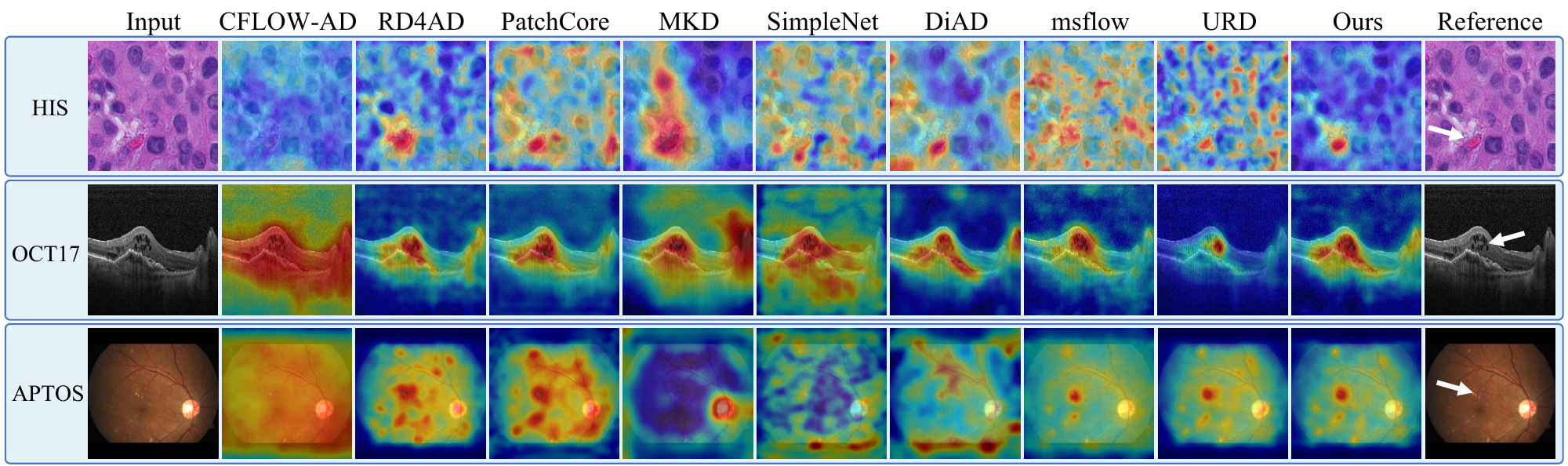}}
\caption{Visualization of anomaly detection maps for classification tasks. From left to right: input images, predictions from CFLOW-AD, RD4AD, PatchCore, MKD, SimpleNet, DiAD, msflow, URD, our method, and reference regions. The rows show examples from histopathology (HIS), OCT17, and APTOS, respectively. Reference regions are provided only for qualitative visualization and are not used for localization evaluation.}
\label{fig:detection}
\end{figure*}

\begin{figure*}[h!]
\centerline{\includegraphics[width=0.9\textwidth, trim={7cm 6.5cm 5cm 1.8cm}, clip]{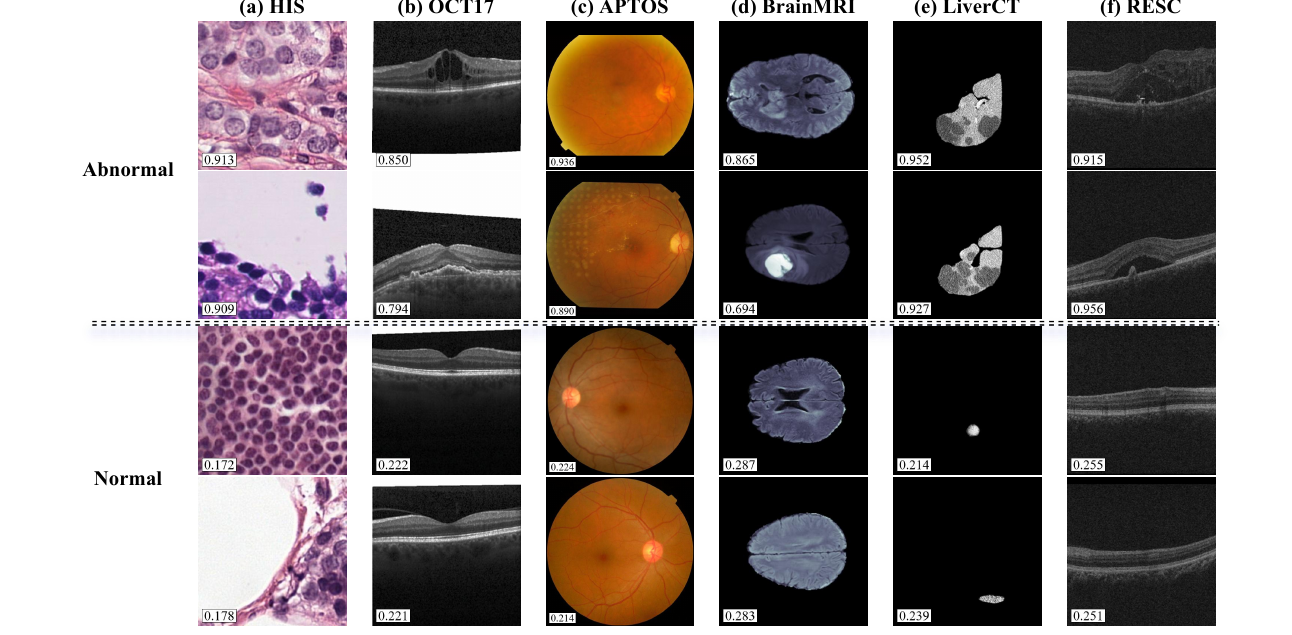}}
\caption{Examples of normal and abnormal samples with predicted anomaly scores across six datasets. Higher scores indicate a higher likelihood of anomaly.}
\label{fig:score_visualization}
\end{figure*}

The advantage is particularly evident in retinal and lesion localization tasks. On OCT17 and APTOS, CFR-Net improves the classification AUROC over the second-best methods by 1.07 and 1.17 percentage points, respectively, showing stable discrimination under layer-wise retinal structures, illumination variation, and lesions with different spatial extents. For localization, CFR-Net surpasses the second-best results by 1.80 percentage points on BrainMRI and 1.83 percentage points on RESC, indicating that the model can better project anomaly evidence back to spatially coherent regions. On LiverCT, CFR-Net also achieves favorable performance in both classification and localization, with AUROC scores of 74.62\% and 98.48\%, respectively.

Taken together, CFR-Net attains a mean AUROC of 87.23\% for classification and 98.61\% for localization, showing particular effectiveness for localization-oriented medical anomaly detection, where shared conditioning before decoding and crossed consistency after decoding jointly support stable normal-pattern reconstruction. At the same time, MKD~\cite{Salehi_2021_CVPR} and PatchCore~\cite{roth2022towards} achieve higher AUROC on HIS and BrainMRI classification, respectively, likely due to multi-resolution distillation for texture-rich pathology patches and memory-based nearest-neighbor aggregation for image-level scoring.

Visual comparisons further support the quantitative results. In Fig.~\ref{fig:localization}, CFR-Net produces compact responses around tumor regions in BrainMRI, better follows layer-wise abnormal structures in RESC, and suppresses vessel-like false positives in LiverCT. Although PatchCore~\cite{roth2022towards} achieves higher AUROC for BrainMRI classification, CFR-Net provides more spatially coherent localization maps, suggesting that its advantage lies in generating interpretable anomaly responses rather than only aggregating local deviations for image-level scoring. These observations are consistent with the role of shared conditioning and crossed consistency in preserving spatially meaningful anomaly responses.

Fig.~\ref{fig:detection} further shows anomaly response maps for classification-only datasets. CFR-Net highlights malignant nuclei clusters in HIS, drusen-related changes in OCT17, and lesion-related regions in APTOS despite focus and illumination variations, complementing the image-level classification results with spatial evidence of pathology-related responses. In Fig.~\ref{fig:score_visualization}, we further provide score-level examples across all evaluated datasets, where normal samples generally receive lower anomaly scores and abnormal samples tend to produce higher responses. The score separation is consistent across different modalities, suggesting that CFR-Net can translate spatial anomaly responses into stable image-level decisions rather than relying on dataset-specific visual cues.

The performance difference between classification and localization reflects their different requirements: classification relies more on image-level score aggregation, whereas localization requires spatially precise and structurally coherent anomaly responses. CFR-Net shows stronger relative advantages in localization, indicating that its coupled design is particularly effective when anomaly evidence needs to be preserved at the feature-map level rather than only aggregated into an image-level score.

\subsection{Ablation Study}

For the ablation studies, AC denotes the average AUROC over all six anomaly classification tasks, including HIS, OCT17, APTOS, BrainMRI, LiverCT, and RESC, while AL denotes the average AUROC over the three anomaly localization tasks, including BrainMRI, LiverCT, and RESC. 

\begin{figure*}[h!]
\centerline{\includegraphics[width=\textwidth, trim={0cm 0cm 0cm 0cm}, clip]{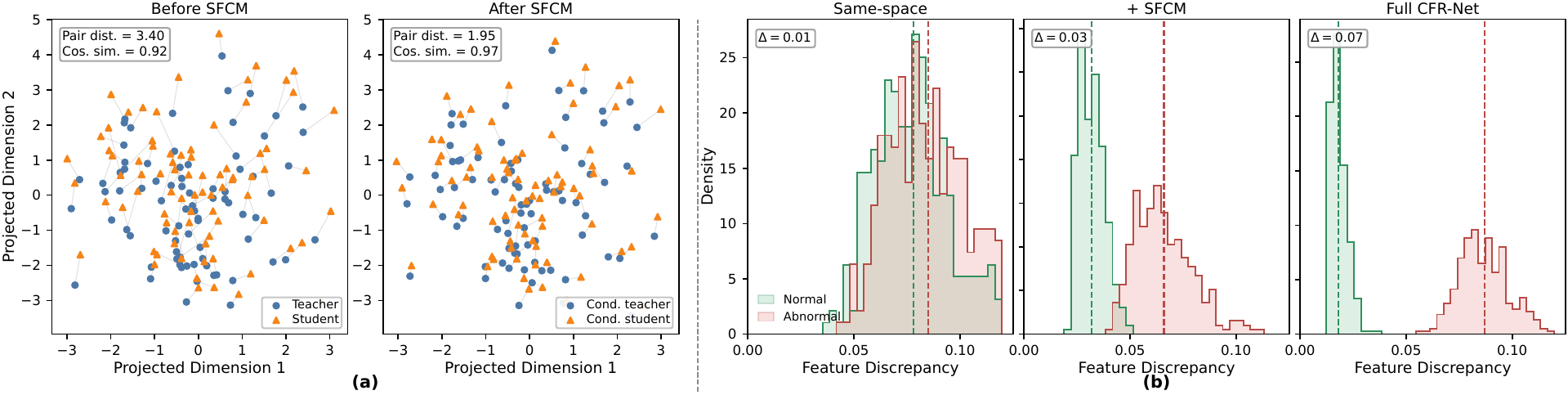}}
\caption{Teacher-student mismatch and discrepancy analysis.
(a) Paired teacher-student representations before and after SFCM. Each line connects the same normal sample in the projected space, and SFCM reduces the paired offset.
(b) Normal and abnormal discrepancy distributions under different matching designs. SFCM reduces normal discrepancies, while the full CFR-Net achieves the largest normal-abnormal separation.}
\label{fig:representation_discrepancy}
\end{figure*}

\textbf{Progressive component analysis.}
Table~\ref{tab:abl_component} presents the progressive ablation of CFR-Net. Introducing the student encoder with same-space matching raises the localization AUROC from 77.37\% to 85.92\%, indicating that a trainable representation space provides useful adaptation beyond the frozen teacher features. Shared feature conditioning brings a further 7.36\% gain in localization AUROC, reaching 93.28\%, by improving the compatibility between the teacher and student representations before decoding, which is also supported by the representation analysis in Fig.~\ref{fig:representation_discrepancy}.

\begin{table}[h!]
\centering
\caption{Progressive component ablation. AC denotes the average AUROC over all six anomaly classification tasks, and AL denotes the average AUROC over the three anomaly localization tasks. \(L_{\mathrm{w}}\) denotes the hierarchical discrepancy weighting term, and \(\mathcal{X}_{\mathrm{hw}}\) denotes the homework subset used in homework set reorganization strategy.}
\label{tab:abl_component}
\small
\setlength{\tabcolsep}{2.3pt}
\begin{tabular}{lccccccc}
\toprule
Variant & \(F_{\mathrm{se}}\) & Match & SFCM & \(L_{\mathrm{w}}\) & \(\mathcal{X}_{\mathrm{hw}}\) & AC & AL \\
\midrule
Backbone only &  & -- &  &  &  & 74.60 & 77.37 \\
+ Student enc. stream & \checkmark & Same &  &  &  & 76.33 & 85.92 \\
+ SFCM & \checkmark & Same & \checkmark &  &  & 84.65 & 93.28 \\
+ Cross-space & \checkmark & Cross & \checkmark &  &  & 85.99 & 94.18 \\
+ \(L_{\mathrm{w}}\) & \checkmark & Cross & \checkmark & \checkmark &  & 86.71 & 96.82 \\
Full CFR-Net & \checkmark & Cross & \checkmark & \checkmark & \checkmark & 87.23 & 98.61 \\
\bottomrule
\end{tabular}
\end{table}

Replacing same-space matching with cross-space consistency yields additional gains of 1.34 and 0.90\% in classification and localization AUROC, respectively, indicating that constraining each decoded stream with the complementary encoder representation is more effective than maintaining stream-specific same-space consistency. Hierarchical discrepancy weighting further brings the two averages to 86.71\% and 96.82\%. Finally, homework-set reorganization adds another 0.52 and 1.79\%, resulting in the final classification and localization AUROCs of 87.23\% and 98.61\%, respectively, showing an additional benefit from periodically reorganizing the normal samples during training. The first five variants use conventional 100-epoch full-set training, whereas the full CFR-Net uses 20 reorganization cycles with a $5+1$ schedule and $\rho=0.5$, corresponding to 60 full-set-equivalent sample passes, which achieves the best performance with a lower overall sample exposure.

\textbf{Teacher-student collaboration design.}
In Table~\ref{tab:abl_shared}, we analyze the effect of parameter sharing in SFCM under the final CFR-Net framework. Shared SFCM achieves higher AUROC on average for both classification and localization while using fewer parameters than independent SFCM, indicating that common conditioning rules reduce unnecessary stream-specific transformations and encourage teacher and student streams to organize medical-aware features in a compatible manner.

\begin{table}[h!]
\centering
\caption{Ablation of independent and shared feature conditioning under the full setting.}
\label{tab:abl_shared}
\small
\setlength{\tabcolsep}{5pt}
\begin{tabular}{lccc}
\toprule
Conditioning design & SFCM Params. (M) & AC & AL \\
\midrule
Independent SFCM & 105.84 & 84.96 & 95.45 \\
Shared SFCM & 52.92 & 87.23 & 98.61 \\
\bottomrule
\end{tabular}
\end{table}

Based on the shared conditioning design, we compare different teacher-student interaction patterns under the final CFR-Net framework in Table~\ref{tab:abl_interaction}, where only the matching strategy is changed. Cross-space matching achieves the highest average AUROC for both classification and localization, outperforming same-space matching and direct feature merge. This result, together with the discrepancy analysis in Fig.~\ref{fig:representation_discrepancy}, shows that complementary matching not only improves the final detection performance but also produces a larger separation between normal and abnormal discrepancy responses.
\begin{table}[h!]
\centering
\caption{Ablation of teacher-student interaction patterns under the full setting.}
\label{tab:abl_interaction}
\small
\setlength{\tabcolsep}{5pt}
\begin{tabular}{lcc}
\toprule
Interaction pattern & AC & AL \\
\midrule
Same-space matching & 84.41 & 96.72 \\
Direct feature merge & 82.73 & 95.18 \\
Cross-space matching & 87.23 & 98.61 \\
\bottomrule
\end{tabular}
\end{table}

\textbf{Representation and discrepancy analysis.}
To examine the mismatch between frozen teacher priors and student representations adapted to medical images, we analyze paired representation geometry and feature discrepancy distributions on BrainMRI in Fig.~\ref{fig:representation_discrepancy}. For Fig.~\ref{fig:representation_discrepancy}(a), features from the deepest cross-space level of the fully trained CFR-Net are extracted from all normal test samples, spatially pooled, jointly standardized, and projected into a shared two-dimensional space using Principal Component Analysis (PCA), with 80 paired samples shown for visualization. Before SFCM, the teacher and student representations exhibit noticeable spatial offsets in the shared PCA space despite relatively high directional similarity, with a mean pair distance of 3.40 and a cosine similarity of 0.92. After shared conditioning, the distance decreases to 1.95 while the cosine similarity further increases, indicating improved representation compatibility in both feature proximity and directional consistency.

Fig.~\ref{fig:representation_discrepancy}(b) further shows substantial overlap between normal and abnormal discrepancy distributions under same-space matching. Introducing SFCM shifts normal discrepancies toward lower values and increases the mean normal-abnormal discrepancy gap from 0.01 to 0.03, while cross-space consistency further enlarges it to 0.07 in the full CFR-Net. The discrepancy gaps in Fig.~\ref{fig:representation_discrepancy}(b) measure normal-abnormal separation and are therefore not directly comparable in magnitude to the pairwise similarity metrics in Fig.~\ref{fig:representation_discrepancy}(a). Overall, shared conditioning reduces teacher-student representation mismatch, while complementary cross-space matching improves the separation between normal and abnormal discrepancy responses.

\textbf{Efficiency analysis.}
In Table~\ref{tab:efficiency}, we compare inference efficiency on BrainMRI under a unified setting with the same input resolution, batch size, and A100 GPU. CFR-Net achieves 57.68 FPS with a latency of 17.34 ms, while requiring higher computation than lightweight distillation-based methods due to shared conditioning and dual-stream decoding. It remains substantially more efficient than the diffusion-based DiAD, while the w/o SFCM variant reaches 94.00 FPS with 75.24 G FLOPs, indicating that shared feature conditioning accounts for a major part of the additional computational cost.

\begin{table}[h!]
\centering
\caption{Efficiency comparison. Params. (M), FLOPs (G), FPS, and Latency (ms) denote the number of parameters, floating point operations, frames per second, and per-image inference latency, respectively.}
\label{tab:efficiency}
\small
\setlength{\tabcolsep}{4pt}
\begin{tabular}{lcccc}
\toprule
Method & Params. (M) & FLOPs (G) & FPS & Latency (ms)\\
\midrule
CFLOW-AD & 236.56 & 28.66 & 62.88 & 15.90 \\
RD4AD & 161.08 & 38.85 & 77.39 & 12.92 \\
PatchCore & 68.88 & 12.07 & 38.00 & 26.32 \\
SimpleNet & 72.82 & 18.95 & 4.57 & 218.61 \\
DiAD & 1331.30 & 451.50 & 2.02 & 496.13 \\
URD & 208.93 & 50.82 & 115.44 & 8.66 \\
w/o SFCM & 144.68 & 75.24 & 94.00 & 10.64 \\
CFR-Net & 205.84 & 126.39 & 57.68 & 17.34 \\
\bottomrule
\end{tabular}
\end{table}

\section{Conclusion}
CFR-Net is introduced as a cross-space normality learning framework built upon teacher-student distillation for medical image anomaly detection. It learns normality through shared medical-aware conditioning before decoding and complementary cross-space matching after decoding, improving compatibility between generic teacher priors and student representations adapted to medical images, with the homework set reorganization strategy further stabilizing normality learning by periodically refreshing normal training subsets. Experiments across multiple medical imaging benchmarks demonstrate strong anomaly classification and localization performance. Future work will further investigate image-level anomaly score aggregation to improve performance in classification-oriented scenarios.

\appendix

\subsection{Backbone Analysis}
\begin{table*}[h!]
  \centering
  \small
  \setlength{\tabcolsep}{2.5pt}
  \caption{Comparison with different pretrained encoders across all evaluated datasets. Avg denotes the mean AUROC over all reported classification and localization results.}
  \label{tab:abl_backbone}
  \begin{tabularx}{\textwidth}{@{}c *{10}{>{\centering\arraybackslash}X}@{}}
    \toprule
    \multirow{2}{*}[-2pt]{Backbone} 
& HIS 
& \makebox[\linewidth][c]{OCT17}
& \makebox[\linewidth][c]{APTOS}
& \multicolumn{2}{c}{BrainMRI} 
& \multicolumn{2}{c}{LiverCT} 
& \multicolumn{2}{c}{RESC} 
& \multirow{2}{*}[-2pt]{Avg} \\
    \cmidrule(r){2-2} \cmidrule(lr){3-3} \cmidrule(lr){4-4} 
    \cmidrule(lr){5-6} \cmidrule(lr){7-8} \cmidrule(lr){9-10}
    & AC & AC & AC & AC & AL & AC & AL & AC & AL & \\
    \midrule
    DenseNet121 & 64.22 & 93.12 & 89.08 & 88.01 & 92.90 & 70.65 & 96.38 & 90.21 & 96.77 & 86.82 \\
    EfficientNet-B0 & 62.59 & 95.24 & 91.21 & 88.11 & 95.28 & 69.12 & 95.91 & 89.92 & 97.01 & 87.15 \\
    DeiT-S/16 & 63.58 & 95.83 & 92.25 & 89.25 & 97.16 & 67.01 & 92.58 & 89.20 & 95.20 & 86.90 \\
    ResNet50 & 67.84 & 98.85 & 94.95 & 88.10 & 98.02 & 72.85 & 98.10 & 92.95 & 97.95 & 89.96 \\
    ResNeXt50-32x4d & 68.35 & 99.05 & 95.35 & \textbf{89.96} & 98.15 & 73.20 & 98.18 & 93.05 & 98.02 & 90.37 \\
    ResNet152 & 68.10 & 99.12 & 95.10 & 89.75 & 98.22 & 72.70 & 98.12 & 93.12 & 98.10 & 90.26 \\
    Wide-ResNet101-2 & 67.95 & 98.70 & 94.85 & 88.87 & 98.30 & 73.25 & 98.47 & 92.90 & 98.20 & 90.17 \\
    Wide-ResNet50-2 & \textbf{70.62} & \textbf{99.63} & \textbf{96.07} & 88.42 & \textbf{98.77} & \textbf{74.62} & \textbf{98.48} & \textbf{93.99} & \textbf{98.57} & \textbf{91.02} \\
    \bottomrule
  \end{tabularx}
\end{table*}

In Table~\ref{tab:abl_backbone}, we compare CFR-Net with ResNet-style, non-ResNet convolutional, and transformer-based pretrained encoders. CFR-Net maintains effective performance across these backbones, demonstrating broad compatibility with different encoder families. Wide-ResNet50-2 achieves the strongest overall result with an average AUROC of 91.02\%, where its channel capacity and hierarchical convolutional representations provide a favorable basis for shared conditioning and cross-space consistency.

\subsection{Hyperparameter and Configuration Sensitivity}
\begin{table}[h!]
\centering
\caption{Sensitivity analysis of hierarchical discrepancy weighting.}
\label{tab:hyper_ablation}
\small
\setlength{\tabcolsep}{5pt}
\begin{tabular}{lccc}
\toprule
Parameter & Schedule & AC & AL \\
\midrule
\multirow{4}{*}{\(\alpha\)}
& \(0.10\rightarrow0.05\) & 86.45 & 98.12 \\
& \(0.20\rightarrow0.10\) & 86.81 & 98.36 \\
& \(0.30\rightarrow0.20\) & \textbf{87.23} & \textbf{98.61} \\
& \(0.40\rightarrow0.30\) & 86.92 & 98.43 \\
\midrule
\multirow{4}{*}{\(\beta\)}
& \(0.05\rightarrow0.01\) & 86.57 & 98.20 \\
& \(0.10\rightarrow0.05\) & \textbf{87.23} & \textbf{98.61} \\
& \(0.20\rightarrow0.10\) & 86.88 & 98.39 \\
& \(0.30\rightarrow0.20\) & 86.61 & 98.24 \\
\bottomrule
\end{tabular}
\end{table}

We further evaluate the sensitivity of hierarchical discrepancy weighting and SFCM configurations. In Table~\ref{tab:hyper_ablation}, we compare different decay schedules for \(\alpha\) and \(\beta\), where \(\alpha\) controls dispersion-based level weighting and \(\beta\) controls threshold-based amplification. The default schedules, \(\alpha:0.30\rightarrow0.20\) and \(\beta:0.10\rightarrow0.05\), achieve the strongest overall results, while nearby settings remain competitive, indicating that CFR-Net is stable around the default weighting configuration.

\begin{table}[h!]
\centering
\caption{Configuration sensitivity of SFCM on BrainMRI.}
\label{tab:sfcm_config}
\small
\setlength{\tabcolsep}{4pt}
\begin{tabular}{lccc}
\toprule
Component & Configuration & AC & AL \\
\midrule
\multirow{4}{*}{Patch size}
& \(2\times2\) & 87.91 & 98.18 \\
& \(4\times4\) & \textbf{88.42} & \textbf{98.77} \\
& \(8\times8\) & 87.68 & 97.96 \\
& \(16\times16\) & 87.31 & 97.65 \\
\midrule
\multirow{5}{*}{Pyramid scale}
& \((1,3,5)\) & 87.72 & 97.83 \\
& \((3,5,7)\) & \textbf{88.42} & \textbf{98.77} \\
& \((5,7,9)\) & 87.96 & 98.05 \\
& Single scale \((5)\) & 87.31 & 97.18 \\
& \((1,7,13)\) & 88.08 & 98.24 \\
\midrule
\multirow{6}{*}{Asym. kernel}
& \((1\times3,3\times1,3\times3)\) & \textbf{88.42} & \textbf{98.77} \\
& Horizontal only & 87.94 & 98.20 \\
& Vertical only & 87.68 & 98.11 \\
& Square only & 87.82 & 97.96 \\
& Large scale & 87.59 & 97.88 \\
& Mixed scales & 88.05 & 98.32 \\
\bottomrule
\end{tabular}
\end{table}

In Table~\ref{tab:sfcm_config}, we examine the main configuration choices of SFCM on BrainMRI. The default settings, including a \(4\times4\) patch size, \((3,5,7)\) pyramid kernels, and \((1\times3,3\times1,3\times3)\) asymmetric kernels, achieve the strongest results among the evaluated alternatives. These choices match the intended roles of SFCM: local patch-dispersion modeling, moderate multi-scale context aggregation, and direction-sensitive structural modeling.

\begin{figure}[h!]
\centerline{\includegraphics[width=0.8\columnwidth, trim={0cm 0cm 0cm 0cm}, clip]{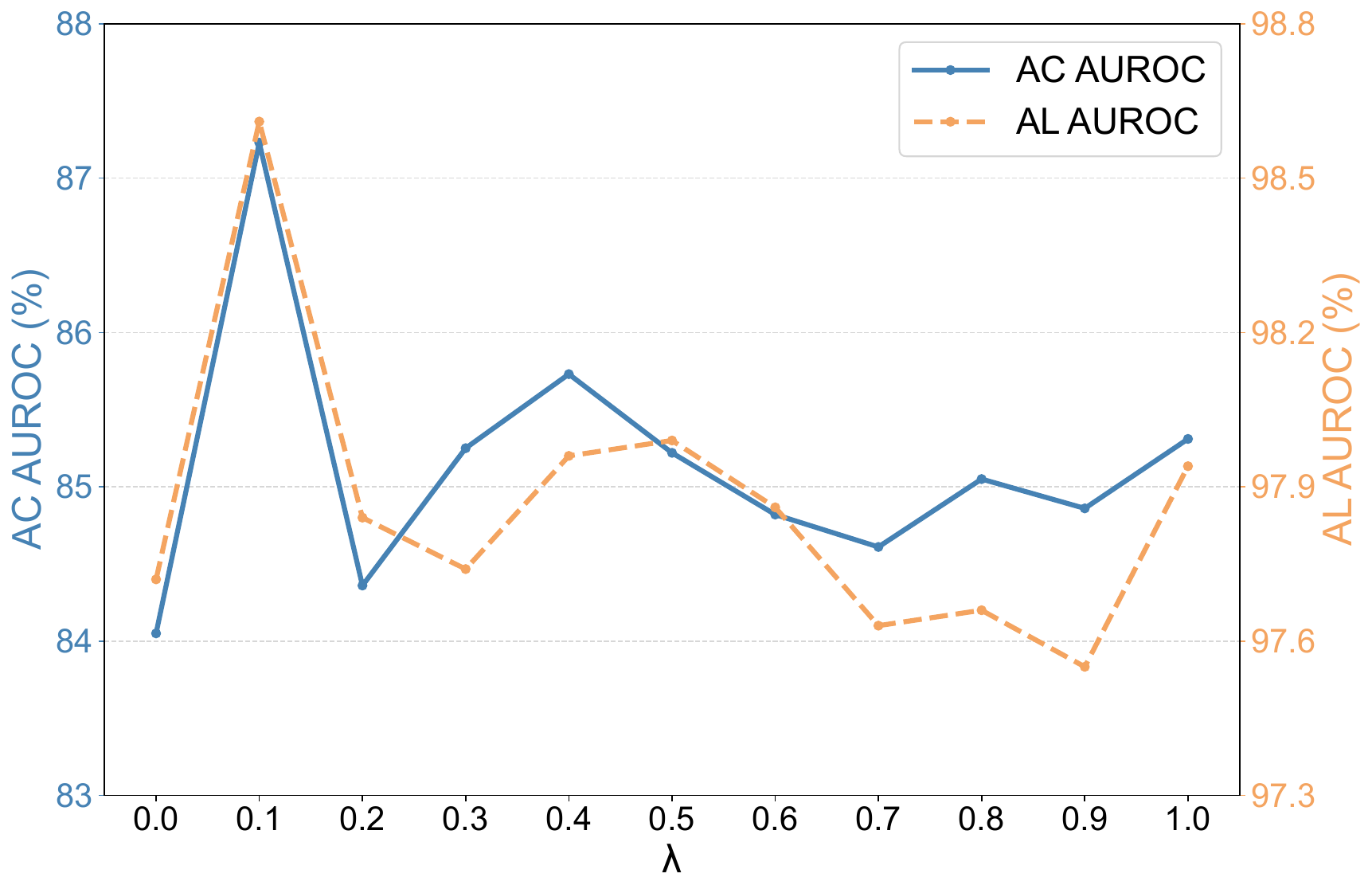}}
\caption{Impact of hyperparameter \(\lambda\).}
\label{fig:lambda}
\end{figure}

The influence of the loss balancing parameter \(\lambda\) is shown in Fig.~\ref{fig:lambda}. The setting \(\lambda=0.1\) achieves the strongest overall performance, suggesting a suitable balance between the weighted and unweighted consistency terms.

\begin{figure}[!h]
\centerline{\includegraphics[width=\columnwidth, trim={0cm 0cm 0cm 0cm}, clip]{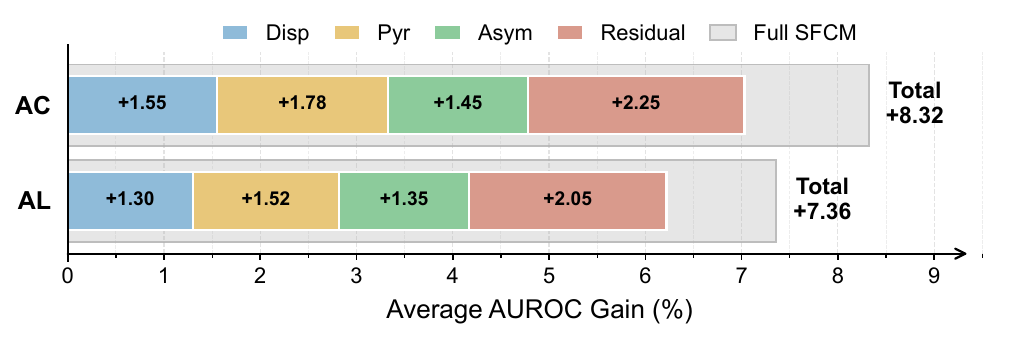}}
\caption{Component contribution analysis of SFCM. Colored segments show the estimated AUROC contribution of each component under a leave-one-component-out protocol. The gray bar denotes the overall gain of the full SFCM over the variant without SFCM. Abbreviations: Disp = Patch-Dispersion Branch, Pyr = Pyramid Context Branch, Asym = Asymmetric Structure Branch, and Residual = residual connection.}
\label{fig:branch}
\end{figure}

\begin{table}[h!]
\centering
\caption{Sensitivity analysis of the homework set reorganization strategy on BrainMRI.}
\label{tab:homework_sensitivity}
\small
\setlength{\tabcolsep}{5.5pt}
\begin{tabular}{c c cc}
\toprule
Parameter & Setting & AC (\%) & AL (\%) \\
\midrule

\multirow{3}{*}{$\rho$}
& 0.25 & 87.91 & 98.54 \\
& 0.50 & \textbf{88.42} & \textbf{98.77} \\
& 0.75 & 88.08 & 98.61 \\
\midrule

\multirow{3}{*}{Schedule}
& $3+1$ & 88.03 & 98.58 \\
& $5+1$ & \textbf{88.42} & \textbf{98.77} \\
& $7+1$ & 88.16 & 98.65 \\
\midrule

\multirow{3}{*}{Cycles}
& 10 & 87.96 & 98.52 \\
& 20 & \textbf{88.42} & \textbf{98.77} \\
& 30 & 88.12 & 98.63 \\
\bottomrule
\end{tabular}
\end{table}
We further examine the configuration sensitivity of the homework set reorganization strategy on BrainMRI. As shown in Table~\ref{tab:homework_sensitivity}, one factor is varied at a time while the remaining settings are fixed at $\rho=0.5$, a $5+1$ schedule, and 20 cycles. Moderate variations in the subset ratio, schedule, and number of cycles lead to limited performance changes, with the default configuration yielding the best overall result. Because these settings also alter the total sample exposure, this analysis evaluates practical configuration sensitivity rather than strictly exposure-matched optimization schedules.

\subsection{Internal SFCM Component Analysis}
We further analyze the internal design of SFCM in Fig.~\ref{fig:branch}. The colored segments estimate the contribution of each component using a leave-one-component-out protocol, where one branch or the residual connection is removed from the full SFCM at a time. The gray bar denotes the overall gain of the full SFCM over the variant without SFCM. Removing any component reduces performance, indicating that the Patch-Dispersion Branch, Pyramid Context Branch, Asymmetric Structure Branch, and residual connection provide complementary contributions to shared feature conditioning. The full SFCM improves the average AUROC by 8.32\% for anomaly classification and 7.36\% for anomaly localization over the variant without SFCM, exceeding the direct leave-one-component contributions and indicating a joint effect among patch-level dispersion modeling, pyramid context aggregation, asymmetric structural modeling, and residual preservation.

\FloatBarrier
\section*{References}

\bibliographystyle{IEEEtran}
\bibliography{reference}

@article{IQBAL2024112650,
title={Multi-scale feature reconstruction network for industrial anomaly detection},
  author={Iqbal, Ehtesham and Khan, Samee Ullah and Javed, Sajid and Moyo, Brain and Zweiri, Yahya and Abdulrahman, Yusra},
  journal={Knowledge-Based Systems},
  volume={305},
  pages={112650},
  year={2024},
  publisher={Elsevier}
}

@article{TONG2023110611,
title = {Two-stage reverse knowledge distillation incorporated and Self-Supervised Masking strategy for industrial anomaly detection},
journal = {Knowledge-Based Systems},
volume = {273},
pages = {110611},
year = {2023},
issn = {0950-7051},
doi = {https://doi.org/10.1016/j.knosys.2023.110611},
url = {https://www.sciencedirect.com/science/article/pii/S0950705123003611},
author = {Guoxiang Tong and Quanquan Li and Yan Song}}

@article{BAI2024112397,
title={Dual-path frequency discriminators for few-shot anomaly detection},
  author={Bai, Yuhu and Zhang, Jiangning and Chen, Zhaofeng and Dong, Yuhang and Cao, Yunkang and Tian, Guanzhong},
  journal={Knowledge-Based Systems},
  volume={302},
  pages={112397},
  year={2024},
  publisher={Elsevier}
}

@article{MA2025113740,
title = {Integrating local and global correlations with Mamba-Transformer for multi-class anomaly detection},
journal = {Knowledge-Based Systems},
volume = {324},
pages = {113740},
year = {2025},
issn = {0950-7051},
doi = {https://doi.org/10.1016/j.knosys.2025.113740},
url = {https://www.sciencedirect.com/science/article/pii/S0950705125007865},
author = {Zeqi Ma and Jiaxing Li and Kaihang Jiang and Wai Keung Wong}}

@inproceedings{liu2025unlocking,
  title={Unlocking the potential of reverse distillation for anomaly detection},
  author={Liu, Xinyue and Wang, Jianyuan and Leng, Biao and Zhang, Shuo},
  booktitle={Proceedings of the AAAI Conference on Artificial Intelligence},
  volume={39},
  number={6},
  pages={5640--5648},
  year={2025}
}

@article{fernando2021deep,
  title={Deep learning for medical anomaly detection--a survey},
  author={Fernando, Tharindu and Gammulle, Harshala and Denman, Simon and Sridharan, Sridha and Fookes, Clinton},
  journal={ACM Computing Surveys (CSUR)},
  volume={54},
  number={7},
  pages={1--37},
  year={2021},
  publisher={ACM New York, NY, USA}
}

@article{shen2017deep,
  title={Deep learning in medical image analysis},
  author={Shen, Dinggang and Wu, Guorong and Suk, Heung-Il},
  journal={Annual review of biomedical engineering},
  volume={19},
  number={1},
  pages={221--248},
  year={2017},
  publisher={Annual Reviews}
}

@inproceedings{wu2022tinyvit,
  title={Tinyvit: Fast pretraining distillation for small vision transformers},
  author={Wu, Kan and Zhang, Jinnian and Peng, Houwen and Liu, Mengchen and Xiao, Bin and Fu, Jianlong and Yuan, Lu},
  booktitle={European conference on computer vision},
  pages={68--85},
  year={2022},
  organization={Springer}
}

@inproceedings{tschuchnig2021anomaly,
  title={Anomaly detection in medical imaging-a mini review},
  author={Tschuchnig, Maximilian E and Gadermayr, Michael},
  booktitle={International Data Science Conference},
  pages={33--38},
  year={2021},
  organization={Springer}
}

@inproceedings{huang2024adapting,
  title={Adapting visual-language models for generalizable anomaly detection in medical images},
  author={Huang, Chaoqin and Jiang, Aofan and Feng, Jinghao and Zhang, Ya and Wang, Xinchao and Wang, Yanfeng},
  booktitle={Proceedings of the IEEE/CVF Conference on Computer Vision and Pattern Recognition},
  pages={11375--11385},
  year={2024}
}

@article{chen2022recent,
  title={Recent advances and clinical applications of deep learning in medical image analysis},
  author={Chen, Xuxin and Wang, Ximin and Zhang, Ke and Fung, Kar-Ming and Thai, Theresa C and Moore, Kathleen and Mannel, Robert S and Liu, Hong and Zheng, Bin and Qiu, Yuchen},
  journal={Medical image analysis},
  volume={79},
  pages={102444},
  year={2022},
  publisher={Elsevier}
}

@article{litjens2017survey,
  title={A survey on deep learning in medical image analysis},
  author={Litjens, Geert and Kooi, Thijs and Bejnordi, Babak Ehteshami and Setio, Arnaud Arindra Adiyoso and Ciompi, Francesco and Ghafoorian, Mohsen and Van Der Laak, Jeroen Awm and Van Ginneken, Bram and S{\'a}nchez, Clara I},
  journal={Medical image analysis},
  volume={42},
  pages={60--88},
  year={2017},
  publisher={Elsevier}
}

@article{aggarwal2021diagnostic,
  title={Diagnostic accuracy of deep learning in medical imaging: a systematic review and meta-analysis},
  author={Aggarwal, Ravi and Sounderajah, Viknesh and Martin, Guy and Ting, Daniel SW and Karthikesalingam, Alan and King, Dominic and Ashrafian, Hutan and Darzi, Ara},
  journal={NPJ digital medicine},
  volume={4},
  number={1},
  pages={65},
  year={2021},
  publisher={Nature Publishing Group UK London}
}

@article{hussein2019lung,
  title={Lung and pancreatic tumor characterization in the deep learning era: novel supervised and unsupervised learning approaches},
  author={Hussein, Sarfaraz and Kandel, Pujan and Bolan, Candice W and Wallace, Michael B and Bagci, Ulas},
  journal={IEEE transactions on medical imaging},
  volume={38},
  number={8},
  pages={1777--1787},
  year={2019},
  publisher={IEEE}
}

@article{guo2023recontrast,
  title={Recontrast: Domain-specific anomaly detection via contrastive reconstruction},
  author={Guo, Jia and Lu, Shuai and Jia, Lize and Zhang, Weihang and Li, Huiqi},
  journal={Advances in Neural Information Processing Systems},
  volume={36},
  pages={10721--10740},
  year={2023}
}

@misc{aptos2019-blindness-detection,
    author = {Karthik and Maggie and Sohier Dane},
    title = {APTOS 2019 Blindness Detection},
    year = {2019},
    howpublished = {\url{https://kaggle.com/competitions/aptos2019-blindness-detection}},
    note = {Kaggle}
}

@misc{baid2021rsnaasnrmiccaibrats2021benchmark,
      title={The rsna-asnr-miccai brats 2021 benchmark on brain tumor segmentation and radiogenomic classification},
  author={Baid, Ujjwal and Ghodasara, Satyam and Mohan, Suyash and Bilello, Michel and Calabrese, Evan and Colak, Errol and Farahani, Keyvan and Kalpathy-Cramer, Jayashree and Kitamura, Felipe C and Pati, Sarthak and others},
  journal={arXiv preprint arXiv:2107.02314},
  year={2021}
}

@article{Bakas2017,
  title={Advancing the cancer genome atlas glioma MRI collections with expert segmentation labels and radiomic features},
  author={Bakas, Spyridon and Akbari, Hamed and Sotiras, Aristeidis and Bilello, Michel and Rozycki, Martin and Kirby, Justin S and Freymann, John B and Farahani, Keyvan and Davatzikos, Christos},
  journal={Scientific data},
  volume={4},
  number={1},
  pages={170117},
  year={2017},
  publisher={Nature Publishing Group}
}

@ARTICLE{6975210,
  title={The multimodal brain tumor image segmentation benchmark (BRATS)},
  author={Menze, Bjoern H and Jakab, Andras and Bauer, Stefan and Kalpathy-Cramer, Jayashree and Farahani, Keyvan and Kirby, Justin and Burren, Yuliya and Porz, Nicole and Slotboom, Johannes and Wiest, Roland and others},
  journal={IEEE transactions on medical imaging},
  volume={34},
  number={10},
  pages={1993--2024},
  year={2014},
  publisher={IEEE}
}

@inproceedings{landman2015miccai,
  title={Miccai multi-atlas labeling beyond the cranial vault--workshop and challenge},
  author={Landman, Bennett and Xu, Zhoubing and Igelsias, Juan and Styner, Martin and Langerak, Thomas and Klein, Arno},
  booktitle={Proc. MICCAI multi-atlas labeling beyond cranial vault—workshop challenge},
  volume={5},
  pages={12},
  year={2015},
  organization={Munich, Germany}
}

@article{BILIC2023102680,
title={The liver tumor segmentation benchmark (lits)},
  author={Bilic, Patrick and Christ, Patrick and Li, Hongwei Bran and Vorontsov, Eugene and Ben-Cohen, Avi and Kaissis, Georgios and Szeskin, Adi and Jacobs, Colin and Mamani, Gabriel Efrain Humpire and Chartrand, Gabriel and others},
  journal={Medical image analysis},
  volume={84},
  pages={102680},
  year={2023},
  publisher={Elsevier}
}

@article{HU2019216,
title={Automated segmentation of macular edema in OCT using deep neural networks},
  author={Hu, Junjie and Chen, Yuanyuan and Yi, Zhang},
  journal={Medical image analysis},
  volume={55},
  pages={216--227},
  year={2019},
  publisher={Elsevier}
}

@article{kermany2018identifying,
  title={Identifying medical diagnoses and treatable diseases by image-based deep learning},
  author={Kermany, Daniel S and Goldbaum, Michael and Cai, Wenjia and Valentim, Carolina CS and Liang, Huiying and Baxter, Sally L and McKeown, Alex and Yang, Ge and Wu, Xiaokang and Yan, Fangbing and others},
  journal={cell},
  volume={172},
  number={5},
  pages={1122--1131},
  year={2018},
  publisher={Elsevier}
}

@article{10.1001/jama.2017.14585,
    title={Diagnostic assessment of deep learning algorithms for detection of lymph node metastases in women with breast cancer},
  author={Ehteshami Bejnordi, Babak and Veta, Mitko and Johannes van Diest, Paul and Van Ginneken, Bram and Karssemeijer, Nico and Litjens, Geert and Van Der Laak, Jeroen AWM and Camelyon16 Consortium and Hermsen, Meyke and Manson, Quirine F and others},
  journal={Jama},
  volume={318},
  number={22},
  pages={2199--2210},
  year={2017}
}

@inproceedings{bao2024bmad,
  title={Bmad: Benchmarks for medical anomaly detection},
  author={Bao, Jinan and Sun, Hanshi and Deng, Hanqiu and He, Yinsheng and Zhang, Zhaoxiang and Li, Xingyu},
  booktitle={Proceedings of the IEEE/CVF Conference on Computer Vision and Pattern Recognition},
  pages={4042--4053},
  year={2024}
}

@InProceedings{Gudovskiy_2022_WACV,
    author    = {Gudovskiy, Denis and Ishizaka, Shun and Kozuka, Kazuki},
    title     = {CFLOW-AD: Real-Time Unsupervised Anomaly Detection With Localization via Conditional Normalizing Flows},
    booktitle = {Proceedings of the IEEE/CVF Winter Conference on Applications of Computer Vision (WACV)},
    month     = {January},
    year      = {2022},
    pages     = {98-107}
}

@InProceedings{Deng_2022_CVPR,
    author    = {Deng, Hanqiu and Li, Xingyu},
    title     = {Anomaly Detection via Reverse Distillation From One-Class Embedding},
    booktitle = {Proceedings of the IEEE/CVF Conference on Computer Vision and Pattern Recognition (CVPR)},
    month     = {June},
    year      = {2022},
    pages     = {9737-9746}
}

@InProceedings{Salehi_2021_CVPR,
    title={Multiresolution knowledge distillation for anomaly detection},
  author={Salehi, Mohammadreza and Sadjadi, Niousha and Baselizadeh, Soroosh and Rohban, Mohammad H and Rabiee, Hamid R},
  booktitle={Proceedings of the IEEE/CVF conference on computer vision and pattern recognition},
  pages={14902--14912},
  year={2021}
}

@inproceedings{liu2023simplenet,
  title={Simplenet: A simple network for image anomaly detection and localization},
  author={Liu, Zhikang and Zhou, Yiming and Xu, Yuansheng and Wang, Zilei},
  booktitle={Proceedings of the IEEE/CVF conference on computer vision and pattern recognition},
  pages={20402--20411},
  year={2023}
}

@inproceedings{he2024diffusion,
  title={A diffusion-based framework for multi-class anomaly detection},
  author={He, Haoyang and Zhang, Jiangning and Chen, Hongxu and Chen, Xuhai and Li, Zhishan and Chen, Xu and Wang, Yabiao and Wang, Chengjie and Xie, Lei},
  booktitle={Proceedings of the AAAI conference on artificial intelligence},
  volume={38},
  number={8},
  pages={8472--8480},
  year={2024}
}

@article{zhou2024msflow,
  title={Msflow: Multiscale flow-based framework for unsupervised anomaly detection},
  author={Zhou, Yixuan and Xu, Xing and Song, Jingkuan and Shen, Fumin and Shen, Heng Tao},
  journal={IEEE Transactions on Neural Networks and Learning Systems},
  year={2024},
  publisher={IEEE}
}

@article{sohn2020learning,
  title={Learning and evaluating representations for deep one-class classification},
  author={Sohn, Kihyuk and Li, Chun-Liang and Yoon, Jinsung and Jin, Minho and Pfister, Tomas},
  journal={arXiv preprint arXiv:2011.02578},
  year={2020}
}

@article{massoli2021mocca,
  title={MOCCA: Multilayer one-class classification for anomaly detection},
  author={Massoli, Fabio Valerio and Falchi, Fabrizio and Kantarci, Alperen and Akti, {\c{S}}eymanur and Ekenel, Hazim Kemal and Amato, Giuseppe},
  journal={IEEE transactions on neural networks and learning systems},
  volume={33},
  number={6},
  pages={2313--2323},
  year={2021},
  publisher={IEEE}
}

@inproceedings{hu2021semantic,
  title={A semantic-enhanced method based on deep SVDD for pixel-wise anomaly detection},
  author={Hu, Chuanfei and Chen, Kai and Shao, Hang},
  booktitle={2021 IEEE International Conference on Multimedia and Expo (ICME)},
  pages={1--6},
  year={2021},
  organization={IEEE}
}

@article{yoa2021self,
  title={Self-supervised learning for anomaly detection with dynamic local augmentation},
  author={Yoa, Seungdong and Lee, Seungjun and Kim, Chiyoon and Kim, Hyunwoo J},
  journal={IEEE Access},
  volume={9},
  pages={147201--147211},
  year={2021},
  publisher={IEEE}
}

@article{bergmann2018improving,
  title={Improving unsupervised defect segmentation by applying structural similarity to autoencoders},
  author={Bergmann, Paul and L{\"o}we, Sindy and Fauser, Michael and Sattlegger, David and Steger, Carsten},
  journal={arXiv preprint arXiv:1807.02011},
  year={2018}
}

@article{chung2020unsupervised,
  title={Unsupervised anomaly detection using style distillation},
  author={Chung, Hwehee and Park, Jongho and Keum, Jongsoo and Ki, Hongdo and Kang, Seokho},
  journal={IEEE Access},
  volume={8},
  pages={221494--221502},
  year={2020},
  publisher={IEEE}
}

@inproceedings{yan2021learning,
  title={Learning semantic context from normal samples for unsupervised anomaly detection},
  author={Yan, Xudong and Zhang, Huaidong and Xu, Xuemiao and Hu, Xiaowei and Heng, Pheng-Ann},
  booktitle={Proceedings of the AAAI conference on artificial intelligence},
  volume={35},
  number={4},
  pages={3110--3118},
  year={2021}
}

@article{song2021anoseg,
  title={AnoSeg: Anomaly segmentation network using self-supervised learning},
  author={Song, Jouwon and Kong, Kyeongbo and Park, Ye-In and Kim, Seong-Gyun and Kang, Suk-Ju},
  journal={arXiv preprint arXiv:2110.03396},
  year={2021}
}

@inproceedings{wyatt2022anoddpm,
  title={Anoddpm: Anomaly detection with denoising diffusion probabilistic models using simplex noise},
  author={Wyatt, Julian and Leach, Adam and Schmon, Sebastian M and Willcocks, Chris G},
  booktitle={Proceedings of the IEEE/CVF conference on computer vision and pattern recognition},
  pages={650--656},
  year={2022}
}

@article{teng2022unsupervised,
  title={Unsupervised visual defect detection with score-based generative model},
  author={Teng, Yapeng and Li, Haoyang and Cai, Fuzhen and Shao, Ming and Xia, Siyu},
  journal={arXiv preprint arXiv:2211.16092},
  year={2022}
}

@inproceedings{bergmann2020uninformed,
  title={Uninformed students: Student-teacher anomaly detection with discriminative latent embeddings},
  author={Bergmann, Paul and Fauser, Michael and Sattlegger, David and Steger, Carsten},
  booktitle={Proceedings of the IEEE/CVF conference on computer vision and pattern recognition},
  pages={4183--4192},
  year={2020}
}

@article{wang2021student,
  title={Student-teacher feature pyramid matching for anomaly detection},
  author={Wang, Guodong and Han, Shumin and Ding, Errui and Huang, Di},
  journal={arXiv preprint arXiv:2103.04257},
  year={2021}
}

@inproceedings{yamada2022reconstructed,
  title={Reconstructed student-teacher and discriminative networks for anomaly detection},
  author={Yamada, Shinji and Kamiya, Satoshi and Hotta, Kazuhiro},
  booktitle={2022 IEEE/RSJ International Conference on Intelligent Robots and Systems (IROS)},
  pages={2725--2732},
  year={2022},
  organization={IEEE}
}

@inproceedings{rudolph2023asymmetric,
  title={Asymmetric student-teacher networks for industrial anomaly detection},
  author={Rudolph, Marco and Wehrbein, Tom and Rosenhahn, Bodo and Wandt, Bastian},
  booktitle={Proceedings of the IEEE/CVF winter conference on applications of computer vision},
  pages={2592--2602},
  year={2023}
}

@article{cao2022informative,
  title={Informative knowledge distillation for image anomaly segmentation},
  author={Cao, Yunkang and Wan, Qian and Shen, Weiming and Gao, Liang},
  journal={Knowledge-Based Systems},
  volume={248},
  pages={108846},
  year={2022},
  publisher={Elsevier}
}

@inproceedings{roth2022towards,
  title={Towards total recall in industrial anomaly detection},
  author={Roth, Karsten and Pemula, Latha and Zepeda, Joaquin and Sch{\"o}lkopf, Bernhard and Brox, Thomas and Gehler, Peter},
  booktitle={Proceedings of the IEEE/CVF conference on computer vision and pattern recognition},
  pages={14318--14328},
  year={2022}
}

@article{nie2026few,
  title={Few-shot medical anomaly detection through centroid consultation back and test-time self-calibration},
  author={Nie, Zihan and Xu, Muhao and Cui, Yuan and Wei, Hua and Yi, Wei and Niu, Sijie and Wan, Yi and Wei, Xunbin and Song, Weiye},
  journal={Pattern Recognition},
  pages={113261},
  year={2026},
  publisher={Elsevier}
}

@article{xu2026beyond,
  title={Beyond Feature Mapping: Dual-Heterogeneous Knowledge Distillation with Mamba for Industrial Anomaly Detection},
  author={Xu, Muhao and Nie, Zihan and Fu, Baochen and Chen, Zhuangzhuang and Li, Zijian and Wei, Hua and Wan, Yi and Song, Weiye},
  journal={Expert Systems with Applications},
  pages={131146},
  year={2026},
  publisher={Elsevier}
}

\end{document}